\PassOptionsToPackage{hidelinks}{hyperref}
\PassOptionsToPackage{numbers, compress}{natbib}

\documentclass{article}

\usepackage[preprint]{corl_2026} 

\usepackage[utf8]{inputenc} 
\usepackage[T1]{fontenc}    
\usepackage{hyperref}       
\usepackage{url}            
\usepackage{booktabs}       
\usepackage{amsfonts}       
\usepackage{nicefrac}       
\usepackage{microtype}      
\usepackage{xcolor}         
\usepackage{enumitem}

\usepackage{algorithm}
\usepackage{algpseudocode}
\algrenewcommand\Require{\State \textbf{Input:}\ }
\algrenewcommand\Ensure{\State \textbf{Output:}\ }
\usepackage{amsmath}
\usepackage{svg}
\usepackage{multirow}
\usepackage[table]{xcolor}
\usepackage{colortbl}
\usepackage{bm}
\usepackage{caption}
\usepackage{subcaption}
\usepackage{siunitx}
\title{Frequency-Guided Action Diffusion via Sub-Frequency Manifold Traversal}

\author{
    Junlin Wang\\
    School of Engineering and Applied Science\\
    University of Pennsylvania\\
    \texttt{wangjl@seas.upenn.edu} \\
}

\begin{document}
\maketitle

\begin{abstract}
    Learning visuomotor policies via behavior cloning typically involves mimicking expert demonstrations collected by human operators. However, natural human demonstrations inherently contain high-frequency noise, such as intermittent jerks, pauses, and action jitter. Training policies to directly imitate these raw trajectories inevitably causes the model to inherit these suboptimal behaviors. This pathology is particularly pronounced in diffusion-based policies, where iterative denoising steps can inadvertently amplify high-frequency artifacts at the expense of meaningful fine-grained details. To address these limitations, we present a novel frequency-based algorithm that enables implicit spectral maneuvering and smooth action generation. Our method, \textbf{Frequency Guidance Operator (FGO)}, steers the generation process of diffusion polices by progressively driving the noisy samples through intermediate sub-frequency manifolds with expanding spectral bands. Validated on 15 robotic manipulation tasks from 5 benchmarks, FGO achieves superior performance in enhancing action smoothness and temporal consistency while preserving the details necessary for successful task execution. Project website: \url{https://henrywjl.github.io/frequency-guidance-operator/}.
\end{abstract}



\section{Introduction}\label{sec:introduction}
Diffusion-based policies \citep{ChiXu2023, ZeZha2024} have recently emerged as a promising approach in behavior cloning due to their remarkable ability to model complex multimodal distributions inherent in diverse behaviors. Unlike conventional methods that learn a direct mapping from observations to actions, diffusion-based approaches frame action prediction as a conditional generative modeling problem and employ an iterative denoising process \citep{HoJai2020} to sample actions from noise. Interestingly, this diffusion denoising process inherently follows a coarse-to-fine generation paradigm in the frequency domain \citep{RisHei2022, FalPan2025}. As isotropic Gaussian noise is injected during the forward process, high-frequency components degrade more rapidly than their low-frequency counterparts, leading the reverse process to reconstruct global structures before fine-grained details. This spectral dynamic conceptually mirrors human decision-making, wherein a high-level intent is formulated before being progressively refined into a precise motion plan.

Despite this inherent frequency hierarchy, standard diffusion policies are typically trained to predict vector fields that map directly to the full-frequency data manifold. Learning this broadband mapping is exceptionally challenging, particularly for complex, highly nonlinear tasks where low-frequency intents and high-frequency details are temporally entangled. This issue is further exacerbated in the behavior cloning paradigm, which heavily relies on high-quality expert demonstrations for supervised learning. In practice, such near-optimal data is rarely accessible, as human demonstrations inevitably contain high-frequency noise and suboptimal, corrective micro-adjustments. Consequently, policies trained across the full-frequency spectrum tend to overfit to these spurious high-frequency variations, causing the robot to execute erratic and jerky motor commands during deployment.


In this work, we propose explicitly steering the reverse denoising process in the time domain while implicitly enforcing a spectral hierarchy in the frequency domain. To this end, we present \textbf{Frequency Guidance Operator (FGO)}, a diffusion guidance mechanism that modulates predicted vector fields using frequency-domain inductive biases. During forward diffusion, FGO trains the model to learn multi-band mappings from noise to sub-frequency data manifolds at various cut-off frequencies. During reverse denoising, instead of forcing noisy samples directly toward the full-frequency data manifold, our method progressively routes action trajectories through a hierarchy of sub-frequency manifolds with expanding spectral bands. By explicitly controlling the cut-off frequencies of the sub-frequency manifolds, our approach implicitly preserves the low-frequency global structure while simultaneously attenuating high-frequency noise during the denoising process. Experimental results demonstrate that FGO significantly improves policy performance across a diverse range of robotic manipulation tasks while yielding highly smooth and temporally consistent action trajectories.


Our contributions are summarized as follows. We propose a novel diffusion guidance paradigm that suppresses high-frequency noise during denoising. We conduct extensive evaluations in both simulated and real-world environments, and demonstrate that our method consistently outperforms its counterparts in both success rate and action smoothness. Finally, we provide comprehensive ablation studies to validate the individual effectiveness of our design choices.

\section{Background}\label{sec:background}

\subsection{Diffusion Policy}
Diffusion policies \citep{ChiXu2023, ZeZha2024} stand for a class of diffusion models \citep{HoJai2020, SonMen2020} that formulate action generation as a conditional iterative denoising process. Specifically, at each time step $t$, the diffusion policy takes a history of $T_o$ observations $\mathbf{O}_t = \{o_{t - T_o + 1}, \dots, o_t\}$ as input and predicts a chunk of $T_a$ actions $\mathbf{A}_t = \{a_t, \dots, a_{t + T_a - 1}\}$. During training, for a diffusion step $k$, isotropic Gaussian noise $\boldsymbol{\epsilon} \sim \mathcal{N}(\mathbf{0}, \mathbf{I})$ is injected to perturb the clean action $\mathbf{A}_t^0$ based on a predefined noise schedule $\alpha_k$ \citep{NicDha2021}:
\begin{equation}\label{eq:forward_diffusion}
    \mathbf{A}_t^k = \sqrt{\bar{\alpha}_k}\mathbf{A}_t^0 + \sqrt{1 - \bar{\alpha}_k}\boldsymbol{\epsilon},
\end{equation}
where $\bar{\alpha}_k = \prod_{i=1}^{k} \alpha_i$. Conditioned on the observation history $\mathbf{O}_t$ and the diffusion step $k$, a noise predictor $\boldsymbol{\epsilon}_{\theta}(\mathbf{A}_t^k, k, \mathbf{O}_t)$ is trained to predict the injected noise by minimizing the following objective:
\begin{equation}
\min_{\theta}
\;
\mathbb{E}_{\mathbf{A}_t^0,\, k,\, \boldsymbol{\epsilon}}
\left[
\left\|
\boldsymbol{\epsilon}_{\theta}(\mathbf{A}_t^k, k, \mathbf{O}_t) - \boldsymbol{\epsilon}
\right\|^2
\right].
\end{equation}
During inference, starting from pure Gaussian noise $\mathbf{A}_t^K$, the policy network performs $K$ iterations of denoising to steer the action trajectory toward the manifold of noise-free actions:
\begin{equation}
    \mathbf{A}_t^{k-1} = \zeta_k (\mathbf{A}_t^k - \gamma_k \boldsymbol{\epsilon}_{\theta}(\mathbf{A}_t^k, k, \mathbf{O}_t)) + \sigma_k \mathcal{N}(\mathbf{0}, \mathbf{I}),
\end{equation}
where the coefficients $\zeta_k$, $\gamma_k$, and $\sigma_k$ are determined by the noise schedule.

\subsection{Discrete Cosine Transform (DCT)}\label{subsec:dct}
Discrete Cosine Transform (DCT) \citep{AhmNat1974} is an orthogonal transformation that decomposes a time-domain signal into a sum of cosine basis functions of varying frequencies. Specifically, consider an action chunk $\mathbf{A} = [a_0, a_1, \dots, a_{N - 1}]^{\top} \in \mathbb{R}^{N \times D}$, where $N$ is the chunk length and $D$ denotes the action dimension. Applying 1D DCT independently to each dimension yields:
\begin{equation}
    \mathcal{C}_i^d = \sum\limits_{n=0}^{N-1} a_n^d \cos \left[ \frac{\pi}{N}(n + \frac{1}{2})i \right], \quad i = 0, 1, \dots, N - 1, \, d = 1, 2, \dots, D,
\end{equation}
where $a_n^d$ is the value of the $d$-th action dimension at time step $n$, and $\mathcal{C}_i^d$ represents the $i$-th DCT coefficient. Here, we define a discrete low-pass filter $\mathcal{L}_f$ with a cut-off frequency $f \le N$ by retaining only the first $f$ frequency components. Mapping this filtered spectrum back to the time domain via the inverse DCT yields the reconstructed action chunk $\hat{\mathbf{A}} = [\hat{a}_0, \hat{a}_1, \dots, \hat{a}_{N - 1}]^{\top}$, computed as:
\begin{equation}
    \hat{a}_n^d = \frac{1}{N}\left( \mathcal{C}_0^d + 2\sum\limits_{i=1}^{f-1}\mathcal{C}_i^d \cos\left[ \frac{\pi}{N}(n + \frac{1}{2})i \right] \right), \quad n = 0, 1, \dots, N - 1, \, d = 1, 2, \dots, D.
\end{equation}


\begin{figure}[htbp]
    \centering
    \includegraphics[width=\textwidth]{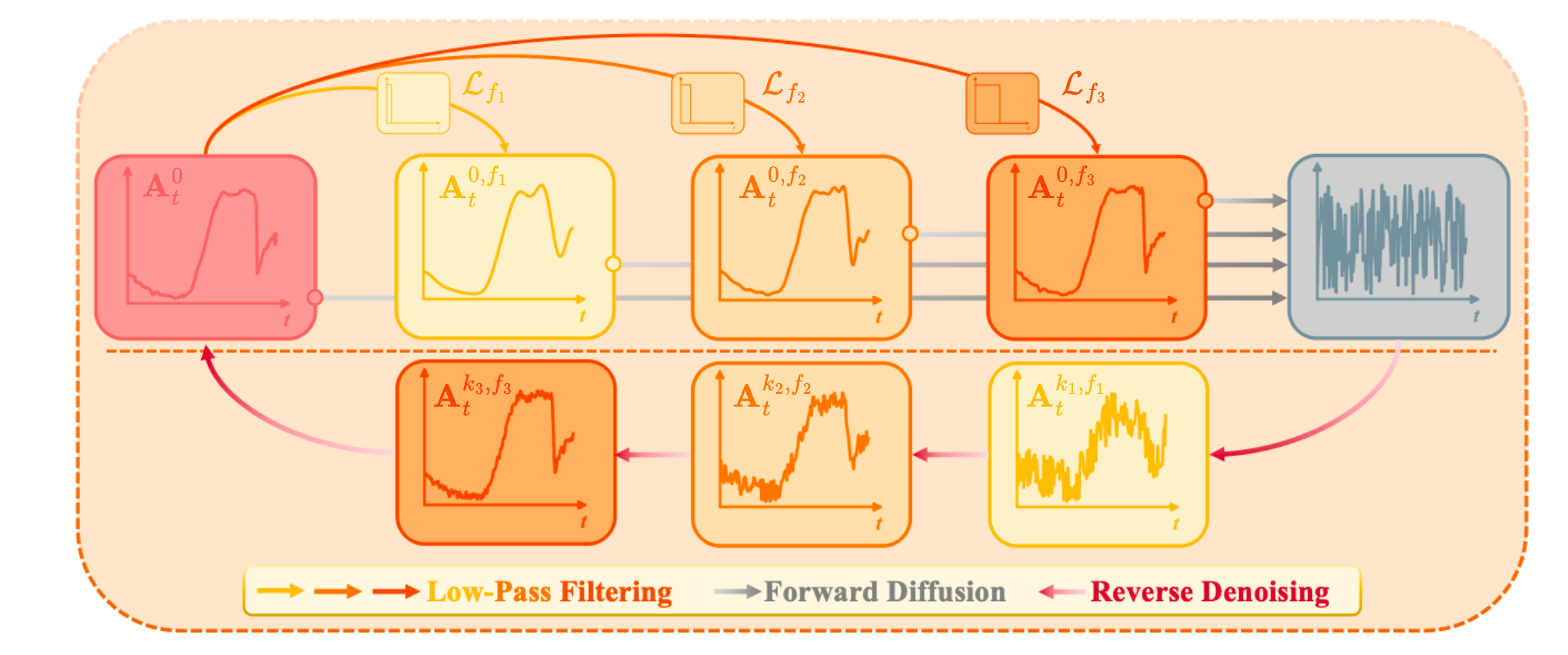}
    \caption{Illustration of FGO. (Top) During the forward diffusion process, full-frequency action trajectories are processed through a bank of low-pass filters, mapping them onto corresponding sub-frequency manifolds. The model is subsequently trained on noise-perturbed variants of these frequency-truncated actions. (Bottom) During the reverse denoising process, the guidance mechanism synthesizes composite vector fields that progressively drives the noisy samples away from the low-frequency foundation and toward the target full-frequency data manifold.}
    \label{fig:fgo-illustration}
\end{figure}

\section{Frequency Guidance Operator (FGO)}\label{sec:methods}

\subsection{Learning Multi-Band Mappings from Noise to Data}

Standard diffusion policies typically learn a full-band mapping directly from noise to full-frequency data manifold. As previously discussed, this broadband objective is significantly challenging and can cause generated samples to drift toward suboptimal trajectories. To address this, we instead propose learning a multi-band mapping that steers noisy samples toward specific sub-frequency manifolds.

Specifically, for a training action chunk $\mathbf{A}_t^0 \in \mathbb{R}^{N \times D}$, we apply the discrete low-pass filter $\mathcal{L}_f$ defined in Section~\ref{subsec:dct} to produce a frequency-truncated sequence $\mathbf{A}_t^{0, f} = \mathcal{L}_f(\mathbf{A}_t^0)$. As demonstrated by \citep{ZhoLiu2025}, utilizing an excessively small cut-off frequency can yield highly distorted inverse DCT reconstructions. To prevent this, we constrain the cut-off frequency $f$ using a hyperparameter $f_{\text{base}} \in [0, N]$, which defines the minimum threshold required to retain the global kinematic structure of the action chunk. Moreover, we explicitly set $f = f_{\text{base}}$ with probability $p_{\text{base}}$, and otherwise sample $f \sim \mathcal{U}(f_{\text{base}}, f_{\text{max}})$, with $f_{\text{max}}$ serving as the spectral upper bound. This sampling strategy establishes a stable baseline essential for our proposed diffusion guidance; we provide ablations for this technique in Section~\ref{subsec:ablation}.

To enable multi-band prediction, we extend the noise predictor to explicitly condition on the cut-off frequency $f$. This yields the augmented parameterization $\boldsymbol{\epsilon}_{\theta}(\mathbf{A}_t^{k, f}, k, \mathbf{O}_t, f)$, where the noisy input $\mathbf{A}_t^{k, f}$ is derived from Equation~(\ref{eq:forward_diffusion}) by substituting the clean action  $\mathbf{A}_t^{0}$ with its frequency-truncated counterpart $\mathbf{A}_t^{0, f}$. By training the model on a spectrum of frequency-truncated action sequences, we empower the policy to selectively target and traverse specific sub-frequency manifolds during inference via frequency conditioning.  The complete training procedure is summarized in Algorithm~\ref{algo:train}.

\subsection{Progressive Guidance Toward the Full-Frequency Manifold}
During the reverse denoising process, we steer the generated trajectory toward the full-frequency data manifold through a series of increasingly complex frequency manifolds. Specifically, at each denoising step $k \in [0, K]$, where $K$ is the total number of diffusion steps, we perform sampling using a linear combination of two conditional noise estimates:
\begin{equation}\label{eq:fgo_guidance}
\tilde{\boldsymbol{\epsilon}} = (1 - \omega_k)\underbrace{\boldsymbol{\epsilon}_{\theta}(\mathbf{A}_t^{k, f_{\text{base}}}, k, \mathbf{O}_t, f_{\text{base}})}_{\boldsymbol{\epsilon}_{\text{base}}} + \omega_k\underbrace{\boldsymbol{\epsilon}_{\theta}(\mathbf{A}_t^{k, f_k}, k, \mathbf{O}_t, f_k)}_{\boldsymbol{\epsilon}_{\text{fine}}}.
\end{equation}
Here, $\boldsymbol{\epsilon}_{\text{base}}$ defines the vector field mapping toward the $f_{\text{base}}$-manifold, while $\boldsymbol{\epsilon}_{\text{fine}}$ defines the vector field mapping toward an intermediate $f_k$-manifold characterized by a higher cut-off frequency $f_k$ ($f_{\text{base}} \leq f_k \leq N$). By interpolating these vector fields via a time-dependent guidance weight $\omega_k$, we explicitly construct a composite vector field that smoothly transitions from a low-frequency manifold to a higher-frequency manifold. As $f_k$ is monotonically increased throughout the reverse process, the sample is systematically propelled through progressively higher-frequency manifolds until it reaches the full-frequency data manifold. In practice, we employ linear schedules for both $f_k$ and $\omega_k$: 
\begin{equation}
f_k = \left\lfloor f_{\text{base}} + (N - f_{\text{base}})\left(1 - \frac{k}{K}\right) \right\rceil, \quad \omega_k = 1 - \frac{k}{K}.
\end{equation}
A remaining technical challenge during inference is acquiring the frequency-truncated noisy inputs $\mathbf{A}_t^{k, f}$. Theoretically, $\mathbf{A}_t^{k, f}$ can be derived from $\mathbf{A}_t^{k}$ and $\mathbf{A}_t^{0}$ (see Appendix~\ref{appendix:derivation} for the full derivation):
\begin{equation}
    \mathbf{A}_t^{k, f} = \mathbf{A}_t^{k} - \sqrt{\bar{\alpha}_k}\mathcal{H}_f(\mathbf{A}_t^{0}),
\end{equation}
where $\mathcal{H}_f$ is a high-pass filter at cut-off frequency $f$. However, this exact formulation is intractable during reverse sampling because the clean action trajectory $\mathbf{A}_t^{0}$ is unknown. As a computationally efficient workaround, we approximate $\mathbf{A}_t^{k, f}$ via direct low-pass filtering of the full-frequency noisy state $\mathbf{A}_t^{k}$, such that $\mathbf{A}_t^{k, f} \approx \mathcal{L}_f(\mathbf{A}_t^{k})$. While this heuristic simultaneously truncates the high-frequency spectra of the injected noise and may introduce minor off-manifold deviations, empirical evaluations confirm its robustness and efficacy. The complete sampling procedure is summarized in Algorithm~\ref{algo:inference}.

\begin{minipage}[t]{0.48\textwidth}
\begin{algorithm}[H]
\caption{Training a Policy with FGO}
\label{algo:train}
\begin{algorithmic}[1]
    \Require Dataset $\mathcal{D} = \{(o_t, a_t)\}_{t=1}^T$, noise predictor $\boldsymbol{\epsilon}_{\theta}$, noise schedule $\alpha_k$, diffusion steps $K$, frequency upper bound $f_{\text{max}}$, base frequency $f_{\text{base}}$, probability $p_{\text{base}}$
    \Repeat
    \State $(\mathbf{O}_t, \mathbf{A}_t^0) \sim \mathcal{D}$
    \State $\begin{cases} 
        f \gets f_{\text{base}} & \text{with prob } p_{\text{base}}, \\
        f \sim \mathcal{U}(f_{\text{base}}, f_{\text{max}}) & \text{otherwise}
    \end{cases}$
    \State $\mathbf{A}_t^{0, f} \gets \mathcal{L}_f(\mathbf{A}_t^0)$
    \State $k \sim \mathcal{U}(0, K)$, $\boldsymbol{\epsilon} \sim \mathcal{N}(\mathbf{0}, \mathbf{I})$
    \State $\mathbf{A}_t^{k, f} \gets \sqrt{\bar{\alpha}_k}\mathbf{A}_t^{0, f} + \sqrt{1 - \bar{\alpha}_k}\boldsymbol{\epsilon}$
    \State $\min\limits_{\theta}\;\mathbb{E}\left[\left\|\boldsymbol{\epsilon}_{\theta}(\mathbf{A}_t^{k, f}, k, \mathbf{O}_t, f) - \boldsymbol{\epsilon}\right\|^2\right]$
    \Until converged
\end{algorithmic}
\end{algorithm}
\end{minipage}
\hfill
\begin{minipage}[t]{0.48\textwidth}
\begin{algorithm}[H]
\caption{Sampling Actions with FGO}
\label{algo:inference}
\begin{algorithmic}[1]
    \Require Observation $\mathbf{O}_t$, noise predictor $\boldsymbol{\epsilon}_{\theta}$, noise schedule $(\zeta_k, \gamma_k, \sigma_k)$, diffusion steps $K$, base frequency $f_{\text{base}}$, frequency schedule $f_k$, guidance weight schedule $\omega_k$
    \State $\hat{\mathbf{A}}_t^K \sim \mathcal{N}(\mathbf{0}, \mathbf{I})$
    \For{$k = K$ \textbf{to} $1$}
        \State $\hat{\mathbf{A}}_t^{k, f_{\text{base}}} \gets \mathcal{L}_{f_{\text{base}}}(\hat{\mathbf{A}}_t^k)$ 
        \State $\hat{\mathbf{A}}_t^{k, f_k} \gets \mathcal{L}_{f_k}(\hat{\mathbf{A}}_t^k)$ 
        
        \State $\boldsymbol{\epsilon}_{\text{base}} \gets \boldsymbol{\epsilon}_{\theta}(\hat{\mathbf{A}}_t^{k, f_{\text{base}}}, k, \mathbf{O}_t, f_{\text{base}})$
        \State $\boldsymbol{\epsilon}_{\text{fine}} \gets \boldsymbol{\epsilon}_{\theta}(\hat{\mathbf{A}}_t^{k, f_k}, k, \mathbf{O}_t, f_k)$
        \State $\tilde{\boldsymbol{\epsilon}} \gets (1 - \omega_k)\boldsymbol{\epsilon}_{\text{base}} + \omega_k\boldsymbol{\epsilon}_{\text{fine}}$
        \State $\hat{\mathbf{A}}_{t}^{k-1} \gets \zeta_k (\hat{\mathbf{A}}_t^k - \gamma_k \tilde{\boldsymbol{\epsilon}}) + \sigma_k \mathcal{N}(\mathbf{0}, \mathbf{I})$
    \EndFor
    \State \Return $\hat{\mathbf{A}}_t^0$
\end{algorithmic}
\end{algorithm}
\end{minipage}

\subsection{$k$-$f$ Coupled (KFC) Sampling}
When sampling the diffusion step $k$ and the cut-off frequency $f$ during policy training, a naive approach is to sample both terms independently. However, this introduces two significant drawbacks. First, since the $f_k$ schedule during inference explicitly dictates that early denoising steps ($k \approx K$) rely exclusively on low-frequency conditions, optimizing the policy network to predict high-frequency manifolds at high noise levels ($k \approx K$) wastes model capacity on unused vector fields. Second, prior work \citep{RisHei2022, FalPan2025} demonstrates that the forward diffusion process degrades high-frequency signals much faster than low-frequency ones. At high noise levels, high-frequency components are entirely dominated by noise, making them particularly difficult to recover during the reverse process.

Motivated by these limitations, we argue that \textit{early denoising steps should not target high-frequency manifolds}. We enforce this constraint during training by dynamically adjusting the upper bound of the cut-off frequency according to the current noise level:
\begin{equation}    
        f_{\text{max}} = \left\lfloor f_{\text{base}} + (N - f_{\text{base}})\left(1 - \frac{k}{K}\right)^{\beta} \right\rceil
\end{equation}
where $\beta \in [0, 1]$ is a hyperparameter controlling the decay rate of the upper bound. When $k$ is small (low noise), the upper bound $f_{\text{max}}$ is high, allowing the model to train across a broad spectrum of frequencies. Conversely, when $k$ is large (high noise), $f_{\text{max}}$ heavily decreases, restricting the model to sample from a narrow band of low frequencies near $f_{\text{base}}$.

\section{Experiments}\label{sec:experiments}
We systematically evaluate FGO on \textbf{15} robotic manipulation tasks from \textbf{5} benchmarks, including four simulation environments and one real-world setup. In the following sections, we outline the baselines and simulation benchmarks, define our evaluation metrics, and present comprehensive experimental findings across both simulation and real-world platforms, alongside detailed ablation studies.
\subsection{Baselines and Simulation Benchmarks}\label{subsec:baselines-simulation-benchmarks}
We compare FGO to the following baselines throughout our experiments:
\begin{enumerate}[label=\arabic*)]
    \item \textbf{3D Diffusion Policy (DP3)} \citep{ZeZha2024}: A CNN-based diffusion policy \citep{ChiXu2023} comprising a lightweight point cloud encoder and a U-Net \citep{RonFis2015} backbone. We include DP3 as a representative baseline and integrate FGO into its architecture by adapting its training and inference pipelines.
    \item \textbf{DiT-Policy} \citep{DasMee2025}: A transformer-based diffusion policy with a Diffusion Transformer (DiT) \citep{PeeXie2023} backbone. We adopt a variant proposed by \citep{ZhuYu2025} and replace the original image encoders with the DP3 encoders \citep{ZeZha2024}.
    \item \textbf{FreqPolicy} \citep{ZhoLiu2025}: A transformer-based autoregressive policy that employs a next-frequency prediction paradigm for action generation. Similar to FGO, it is also trained on multi-band action chunks and progressively recovers the full-frequency predictions during inference.
\end{enumerate}

We perform simulation experiments on \textbf{13} tasks from \textbf{4} established robotic manipulation benchmarks: \textbf{Robosuite} \citep{ZhuWon2020}, \textbf{MimicGen} \citep{ManNas2023}, \textbf{Adroit} \citep{RajKum2017}, and \textbf{DexArt} \citep{BaoXu2023}. Specifically, we select \textbf{6} tasks from Robosuite and MimicGen to evaluate standard parallel-jaw gripper control. We then select \textbf{7} tasks from Adroit and DexArt to evaluate high-dimensional, fine-grained manipulation using two types of dexterous hands. Further details regarding environmental setups are presented in Appendix~\ref{appendix:experimental_setup}.
\subsection{Evaluation Metrics}\label{subsec:evaluation-metrics}

\textbf{Success Rate:} For all benchmark tasks, we report the mean and standard deviation of the success rates for each method across 3 training seeds (0, 1, 2). These results are derived from the best-performing checkpoint, which is evaluated over 50 independent episodes in the simulation environment.

\textbf{Action Total Variation (ATV) \& JerkRMS:} ATV \citep{RudOsh, ParKim2025} measures the temporal consistency of predicted action trajectories by penalizing large step-to-step changes in the control signal. JerkRMS \citep{FlaHog1985, ParKim2025} quantifies the physical smoothness of executed actions by evaluating the root mean square of the motor jerk. Formally, ATV and JerkRMS over a single episode are defined as:
\begin{align}
    \text{ATV} &= \frac{1}{D(T-1)} \sum_{t=1}^{T-1} \sum_{d=1}^{D} \left| a_{t+1}^{d} - a_{t}^{d} \right|, \\
    \text{JerkRMS} &= \sqrt{ \frac{1}{T-1} \sum_{t=1}^{T-1} \| \dddot{\mathbf{q}}_t \|_2^2 },
\end{align}
where $T$ is the episode length, $D$ is the dimensionality of the action space, $a_t^d$ represents the $d$-th component of the motor command at time step $t$, and $\dddot{\mathbf{q}}_t$ denotes the motor jerks at time step $t$.

\textbf{Computational Cost:} We evaluate computational costs across two metrics. \textit{Training time} is reported in GPU hours, measured by training each model for 3,000 epochs with a batch size of 128 on a single NVIDIA RTX 4090 GPU. \textit{Inference speed} is evaluated by measuring the average latency required for a single forward pass of the policy network. 
\begin{table}[htbp]
\caption{Comparison of success rates (\%) on the Robosuite \citep{ZhuWon2020} and MimicGen \citep{ManNas2023} benchmarks. For each task, results are averaged over 3 training seeds and reported as (mean) $\pm$ (standard deviation).}
\label{tab:robosuite-mimicgen}
\centering
\renewcommand\arraystretch{1.5}
\setlength{\tabcolsep}{4pt}
\scalebox{0.75}{
    \begin{tabular}{l !{\hspace{6pt}\vrule\hspace{6pt}} *{4}{S[table-format=3.1(1.1)]} !{\vrule} *{2}{S[table-format=3.1(1.1)]} !{\vrule} S[table-format=1.0]}
    \toprule
    \multirow{2}{*}{\textbf{Algorithms / Tasks}} & \multicolumn{4}{c}{\textbf{Robosuite}} & \multicolumn{2}{c}{\textbf{MimicGen}} & {\multirow{2}{*}{\textbf{Average}}} \\
    \cmidrule(r){2-5}
    \cmidrule(lr){6-7}
     & {Lift} & {Stack} & {Can} & {Square} & {Three Piece Assembly} & {Stack Three} & {} \\
    \midrule
    DP3 \citep{ZeZha2024} & 88.7 \pm 4.2 & 72.0 \pm 2.0 & 64.7 \pm 1.2 & \bfseries 36.7 \pm 1.2 & 35.3 \pm 6.4 & 20.0 \pm 3.5 & 52.9 \\
    DiT-Policy \citep{DasMee2025, ZhuYu2025} & 90.7 \pm 4.2 & 68.7 \pm 7.6 & 64.7 \pm 3.1 & 34.7 \pm 2.3 & 37.3 \pm 7.6 & 18.7 \pm 5.0 & 52.5 \\
    FreqPolicy \citep{ZhoLiu2025} & 89.3 \pm 1.2 & 71.3 \pm 1.2 & 63.3 \pm 2.3 & 36.0 \pm 3.5 & 27.3 \pm 8.1 & 22.0 \pm 4.0 & 51.5 \\
    \rowcolor{orange!20}
    \textbf{FGO (Ours)} & \bfseries 92.7 \pm 3.1 & \bfseries 79.3 \pm 3.1 & \bfseries 66.0 \pm 0.0 & \bfseries 36.7 \pm 3.1 & \bfseries 39.3 \pm 7.0 & \bfseries 25.3 \pm 3.1 & \bfseries 56.6 \\
    \bottomrule
    \end{tabular}
}
\end{table}

\begin{table}[htbp]
\caption{Comparison of success rates (\%) on the Adroit \citep{RajKum2017} and DexArt \citep{BaoXu2023} benchmarks. For each task, results are averaged over 3 training seeds and reported as (mean) $\pm$ (standard deviation).}
\label{tab:dexart-adroit}
\centering
\renewcommand\arraystretch{1.5}
\setlength{\tabcolsep}{4pt}
\scalebox{0.75}{
    \begin{tabular}{l !{\hspace{6pt}\vrule\hspace{6pt}} *{3}{S[table-format=3.1(1.1)]} !{\vrule} *{4}{S[table-format=3.1(1.1)]} !{\vrule} S[table-format=1.0]}
    \toprule
    \multirow{2}{*}{\textbf{Algorithms / Tasks}} & \multicolumn{3}{c}{\textbf{Adroit}} & \multicolumn{4}{c}{\textbf{DexArt}} & {\multirow{2}{*}{\textbf{Average}}} \\
    \cmidrule(r){2-4}
    \cmidrule(lr){5-8}
     & {Hammer} & {Door} & {Pen} & {Laptop} & {Toilet} & {Faucet} & {Bucket} & {} \\
    \midrule
    DP3 \citep{ZeZha2024} & \bfseries 100\pm0.0 & 61.3\pm7.6 & 46.0\pm5.3 & 77.3\pm5.0 & 60.7\pm4.2 & 21.3\pm4.2 & 24.7\pm2.3 & 55.9 \\
    DiT-Policy \citep{DasMee2025, ZhuYu2025} & \bfseries 100\pm0.0 & 63.3\pm7.6 & 52.0\pm2.0 & 75.3\pm3.1 & 63.3\pm5.0 & 20.7\pm3.1 & 19.3\pm3.1 & 56.3 \\
    FreqPolicy \citep{ZhoLiu2025} & 98.7\pm1.2 & 68.0\pm3.5 & 52.0\pm3.5 & 78.0\pm8.0 & 58.7\pm4.6 & 20.7\pm5.0 & 18.7\pm3.1 & 56.4 \\
    \rowcolor{orange!20}
    \textbf{FGO (Ours)} & \bfseries 100\pm0.0 & \bfseries 69.3\pm2.3 & \bfseries 55.3\pm1.2 & \bfseries 81.3\pm6.4 & \bfseries 66.7\pm1.2 & \bfseries 24.0\pm3.5 & \bfseries 25.3\pm2.3 & \bfseries 60.3 \\
    \bottomrule
    \end{tabular}
}
\end{table}
\begin{table}[h!]
\centering
\begin{minipage}{0.45\textwidth}
    \caption{Comparison of ATV and JerkRMS on the Robosuite Can task.}
    \label{tab:atv-jerkrms}
    \renewcommand\arraystretch{1.5}
    \setlength{\tabcolsep}{4pt}
    \scalebox{0.62}{
        \begin{tabular}{l !{\hspace{6pt}\vrule\hspace{6pt}} *{2}{S[table-format=2.2(1.2)]}}
        \toprule
        {Algorithms} & {ATV $\downarrow$ ($\times 10^{-3}$ rad/s)} & {JerkRMS $\downarrow$ (rad/s$^3$)} \\
        \midrule
        DP3 \citep{ZeZha2024} & 14.83 \pm 0.17 & 50.87 \pm 1.27 \\
        DiT-Policy \citep{DasMee2025, ZhuYu2025} & 14.84 \pm 0.22 & 51.01 \pm 1.16 \\
        FreqPolicy \citep{ZhoLiu2025} & 15.25 \pm 0.39 & 46.91 \pm 1.58 \\
        \rowcolor{orange!20}
        \textbf{FGO (Ours)} & \bfseries 14.76 \pm 0.17 & \bfseries 40.79 \pm 0.46 \\
        \bottomrule
        \end{tabular}
    }
\end{minipage}
\hfill
\begin{minipage}{0.5\textwidth}
    \caption{Comparison of training time and inference speed on the Adroit Hammer task.}
    \label{tab:train-inference-time}
    \renewcommand\arraystretch{1.5}
    \setlength{\tabcolsep}{4pt}
    \scalebox{0.62}{
        \begin{tabular}{l !{\hspace{6pt}\vrule\hspace{6pt}} *{2}{S[table-format=2.2]}}
        \toprule
        {Algorithms} & {Training Time $\downarrow$ (GPU h)} & {Inference Speed $\downarrow$ (ms)} \\
        \midrule
        DP3 \citep{ZeZha2024} & 0.47 & 39.49 \\
        DiT-Policy \citep{DasMee2025, ZhuYu2025} & 0.42 & \bfseries 17.20 \\
        FreqPolicy \citep{ZhoLiu2025} & \bfseries 0.35 & 33.49 \\
        \rowcolor{orange!20}
        \textbf{FGO (Ours)} & 0.48 & 44.22 \\
        \bottomrule
        \end{tabular}
    }
\end{minipage}
\end{table}

\subsection{Simulation Benchmark Results}\label{subsec:benchmark-results}
FGO consistently achieves superior or comparable success rates across all tasks when evaluated against the baselines. As detailed in Table~\ref{tab:robosuite-mimicgen}, on the 4 basic visuomotor control tasks selected from the Robosuite benchmark, FGO outperforms all competitors on 3 tasks and maintains comparable performance with DP3 on the remaining one. Moreover, on the 2 more complex MimicGen tasks, which require long-horizon reasoning and fine-grained manipulation, FGO again achieves the highest success rates. As shown in Table~\ref{tab:dexart-adroit}, across the 7 advanced dexterous manipulation tasks involving high-dimensional motor control, FGO surpasses all baseline methods on 6 tasks and yields comparable results on the remaining one. These results demonstrate the robust effectiveness of FGO across different environments and robotic platforms. Additional comparisons are presented in Appendix~\ref{appendix:supplementary-results}.

To evaluate temporal consistency and action smoothness, we select the highly nonlinear Can task from the Robosuite benchmark and compute the ATV and JerkRMS metrics for all methods. For a fair comparison, we follow \cite{ParKim2025} and compute these metrics exclusively over the approach phase (the first 32 time steps) toward the target object, as later trajectories naturally diverge depending on task success or failure. Compared to the baselines, FGO achieves the lowest ATV and JerkRMS scores with a particularly pronounced reduction in JerkRMS. These empirical results demonstrate that FGO effectively enables policies to generate highly smooth and temporally consistent actions.

Finally, we analyze the computational overhead of FGO, as guidance mechanisms naturally incur additional computational demands. As summarized in Table~\ref{tab:train-inference-time}, FGO introduces negligible additional training time compared to the DP3 baseline. During inference, however, FGO exhibits a comparatively higher latency than the baseline methods. This overhead is a well-documented characteristic of guidance-based algorithms and remains a primary direction for future optimization.
\subsection{Real-World Experiments}\label{subsec:real-world-experiments}
\begin{figure}[htbp]
    \centering
    \begin{minipage}[b]{0.66\textwidth}
        \centering
        \begin{subfigure}[b]{0.23\textwidth}
            \centering
            \includegraphics[width=\textwidth]{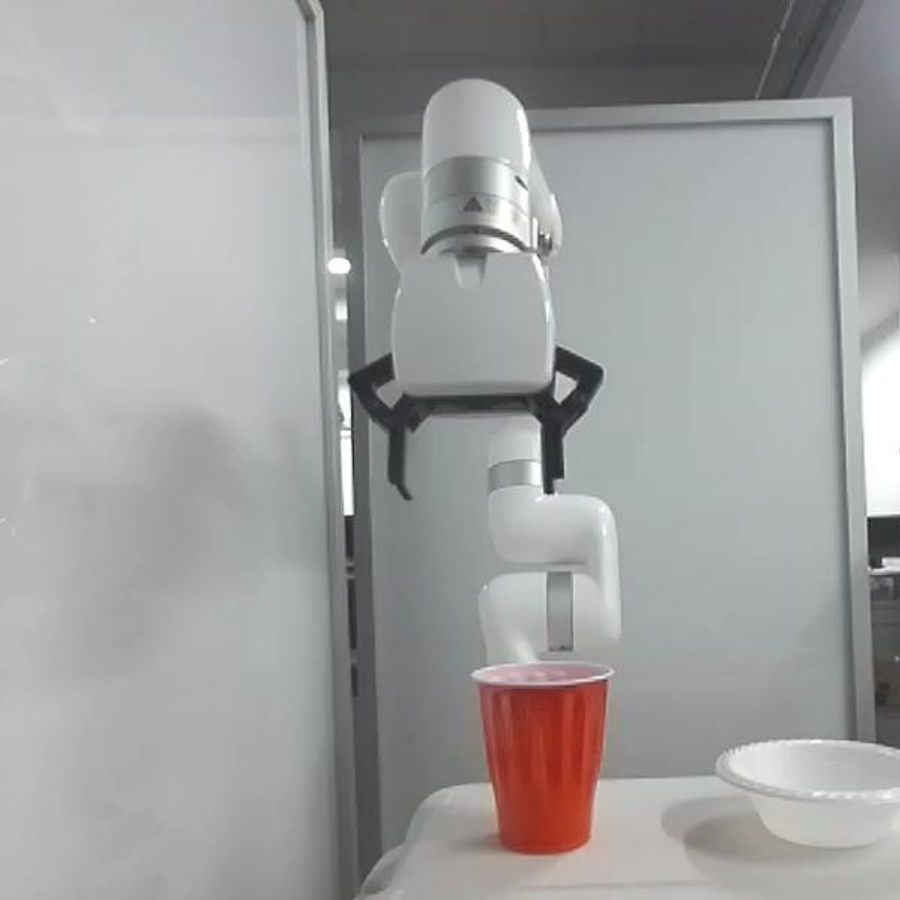}
        \end{subfigure}
        \begin{subfigure}[b]{0.23\textwidth}
            \centering
            \includegraphics[width=\textwidth]{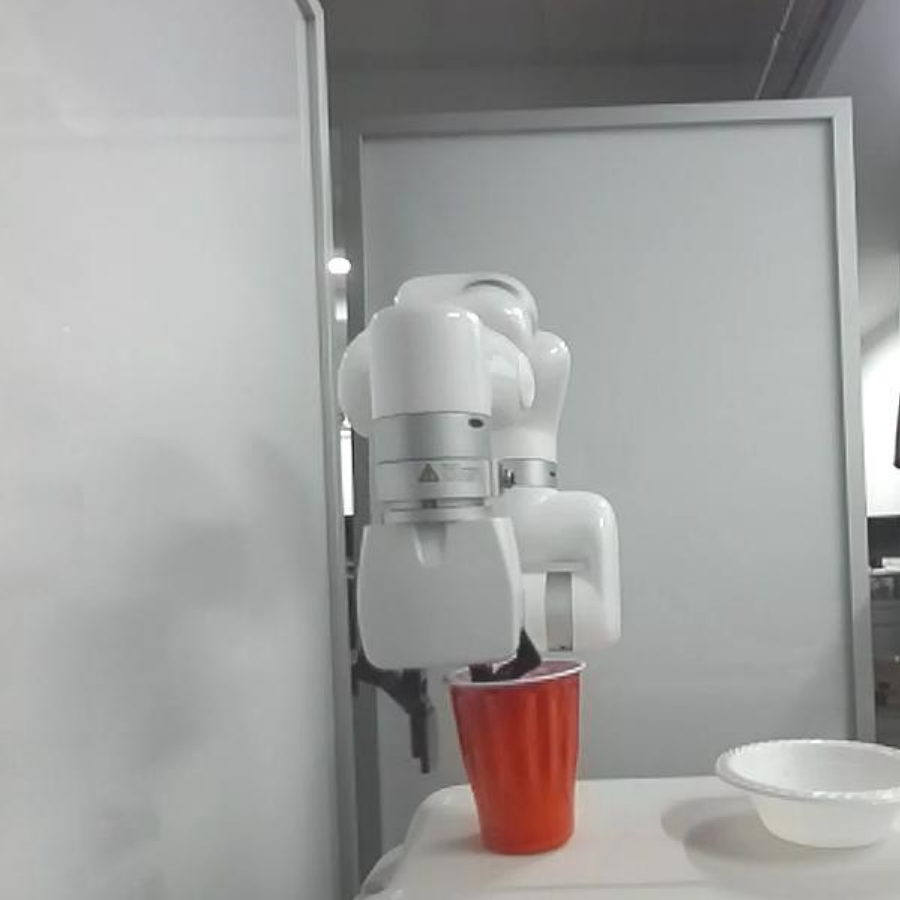}
        \end{subfigure}
        \begin{subfigure}[b]{0.23\textwidth}
            \centering
            \includegraphics[width=\textwidth]{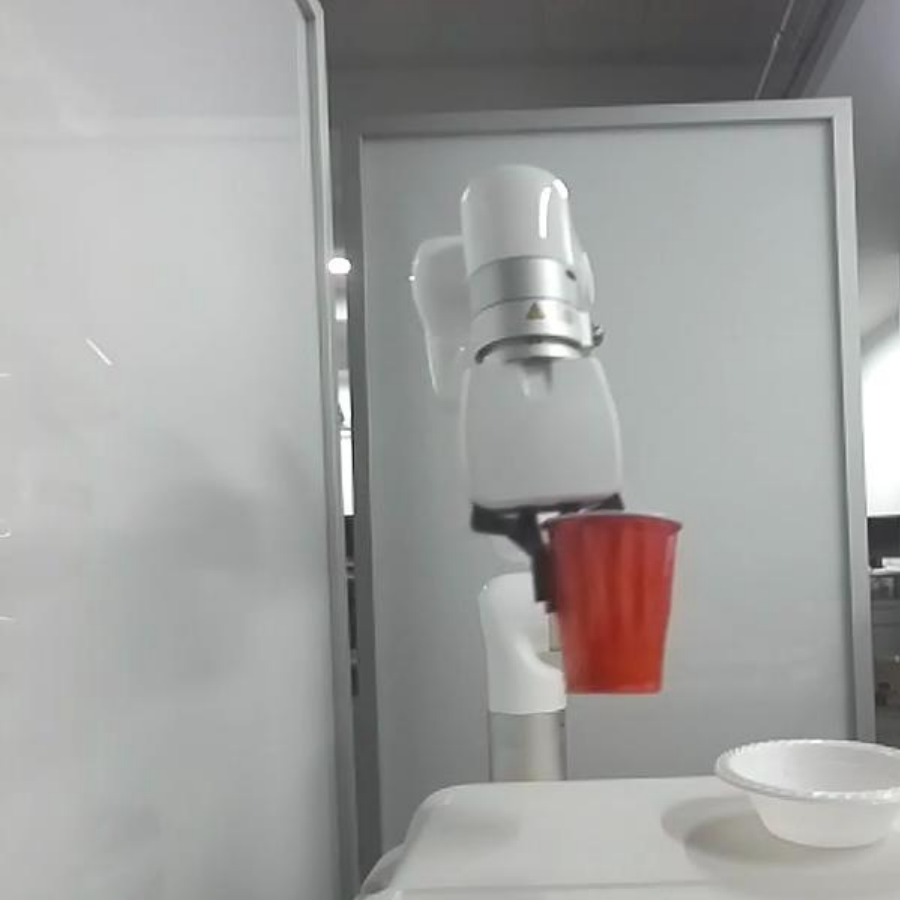}
        \end{subfigure}
        \begin{subfigure}[b]{0.23\textwidth}
            \centering
            \includegraphics[width=\textwidth]{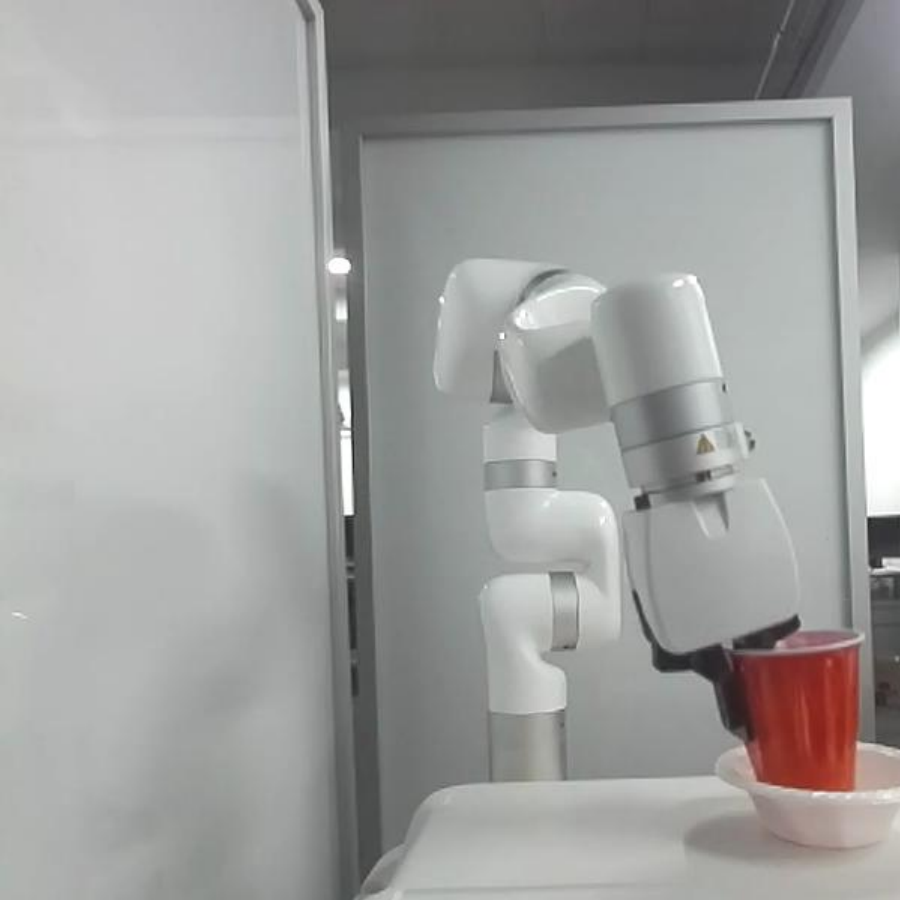}
        \end{subfigure}
        
        \vspace{1mm} 
        
        \begin{subfigure}[b]{0.23\textwidth}
            \centering
            \includegraphics[width=\textwidth]{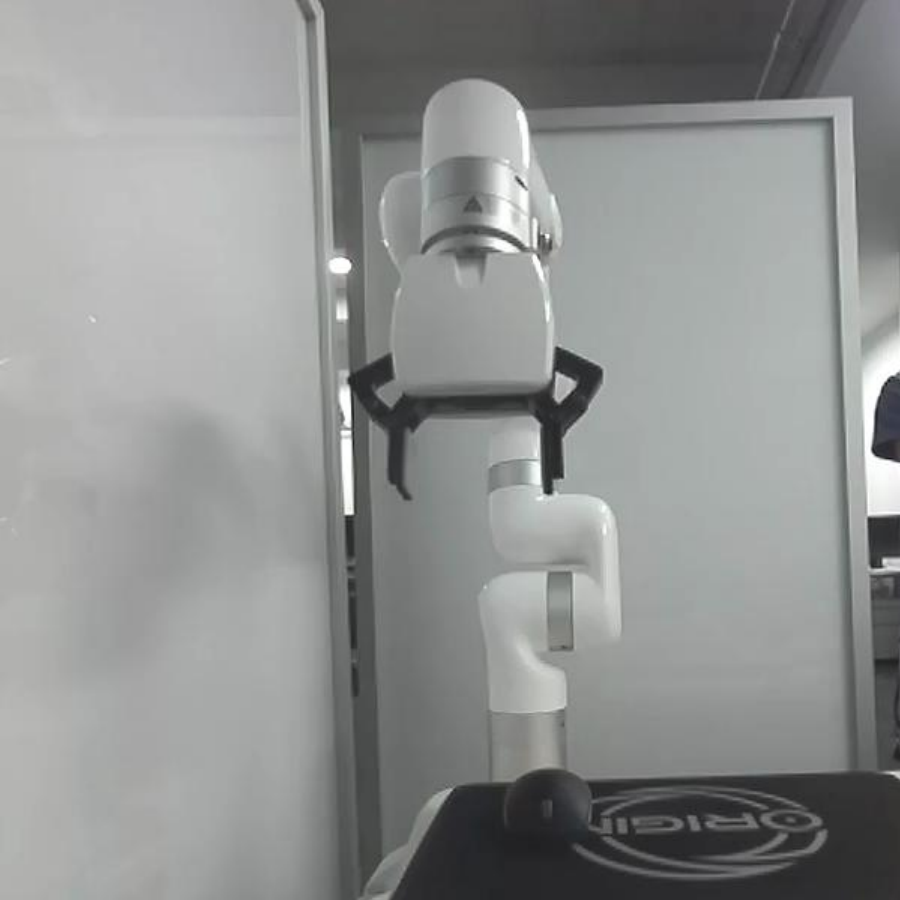}
        \end{subfigure}
        \begin{subfigure}[b]{0.23\textwidth}
            \centering
            \includegraphics[width=\textwidth]{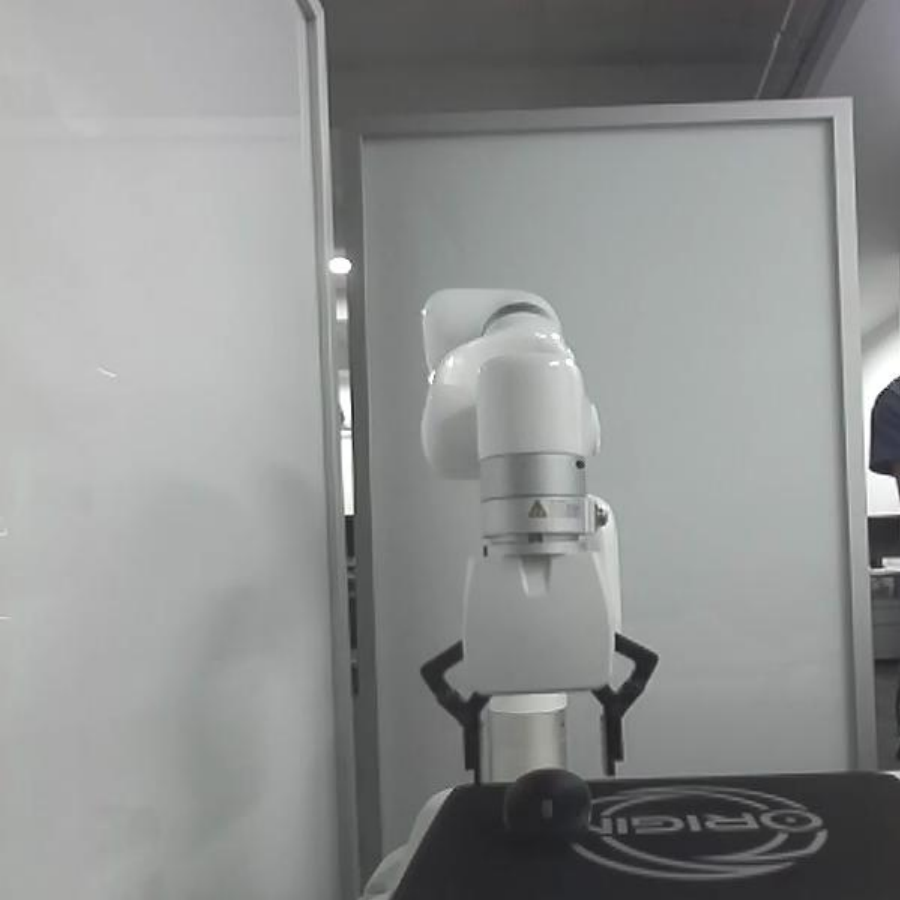}
        \end{subfigure}
        \begin{subfigure}[b]{0.23\textwidth}
            \centering
            \includegraphics[width=\textwidth]{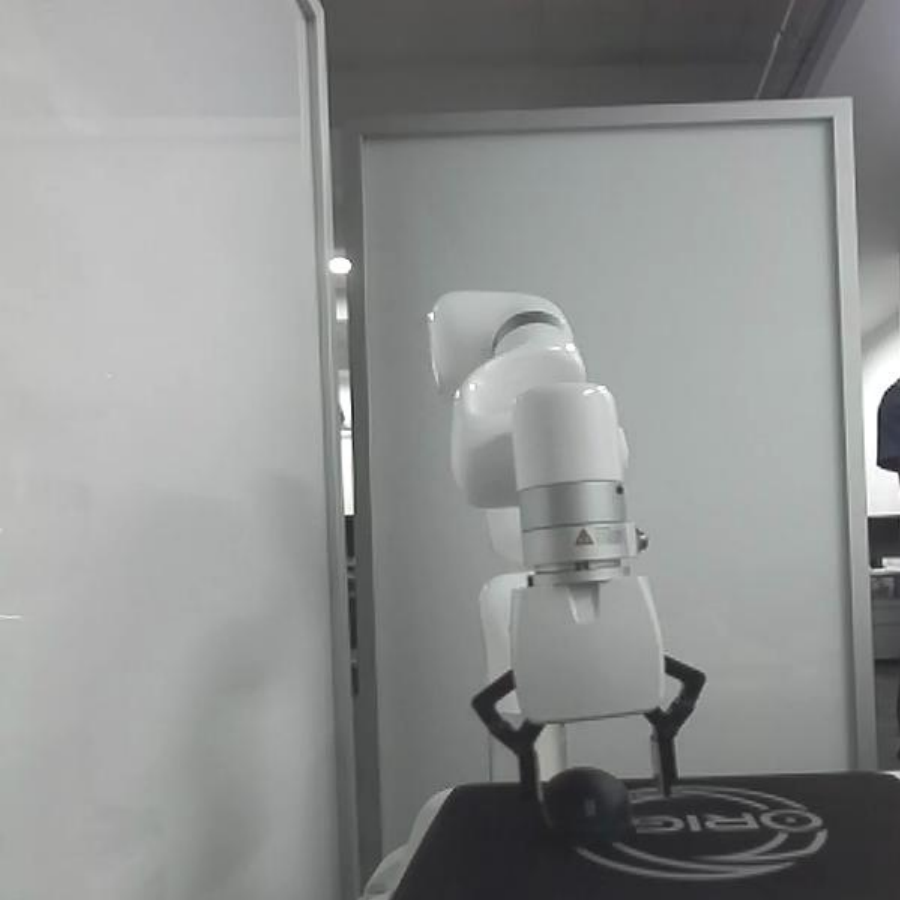}
        \end{subfigure}
        \begin{subfigure}[b]{0.23\textwidth}
            \centering
            \includegraphics[width=\textwidth]{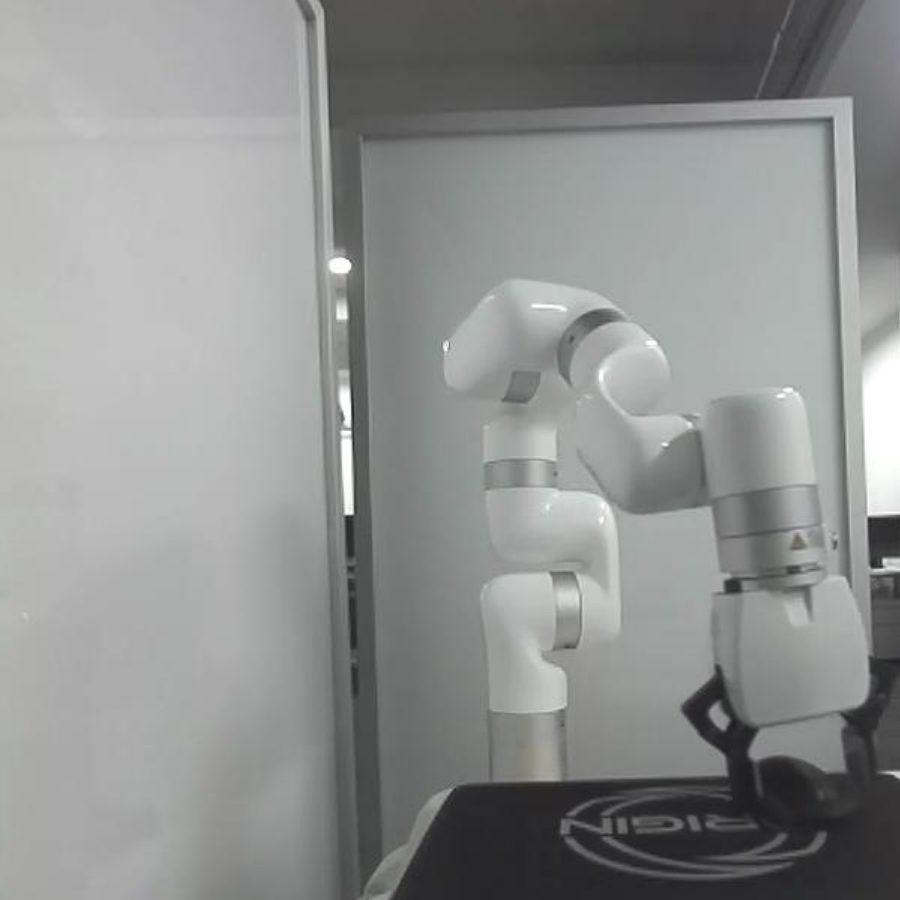}
        \end{subfigure}
    \end{minipage}
    \hfill
    \begin{minipage}[b]{0.3\textwidth}
        \centering
        \begin{subfigure}[b]{\textwidth}
            \centering
            \includegraphics[width=\textwidth]{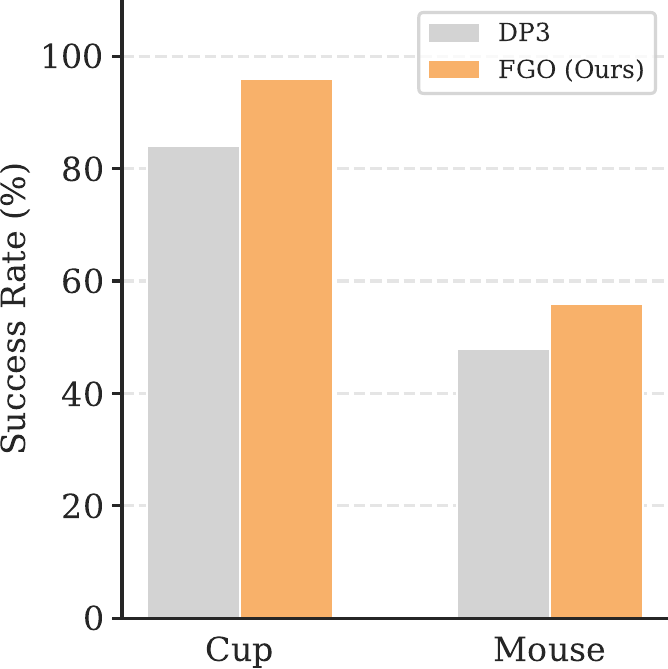}
        \end{subfigure}
    \end{minipage}

    \caption{Real-world experimental setup and results. (Left) Visualizations of the Cup task (top row) and the Mouse task (bottom row) environments. (Right) Success rate comparison on both tasks.}
    \label{fig:real-world-experiments}
\end{figure}

For real-world evaluation, we deploy our policy on an xArm manipulator with a two-finger gripper. As illustrated in Figure~\ref{fig:real-world-experiments} (Left), the evaluation spans two tasks: picking and placing a cup (Cup) and sliding a computer mouse (Mouse). Results in Figure~\ref{fig:real-world-experiments} (Right) demonstrate that FGO consistently outperforms the baseline DP3 method across both tasks, validating its robustness in complex physical environments. Additional details of the experimental setup are provided in Appendices~\ref{appendix:experimental_setup} and \ref{appendix:real-world-experiment-details}.

\subsection{Ablations}\label{subsec:ablation}
\begin{figure}[htbp]
    \centering
    \begin{subfigure}[b]{0.7\textwidth}
        \centering
        \includegraphics[width=\textwidth]{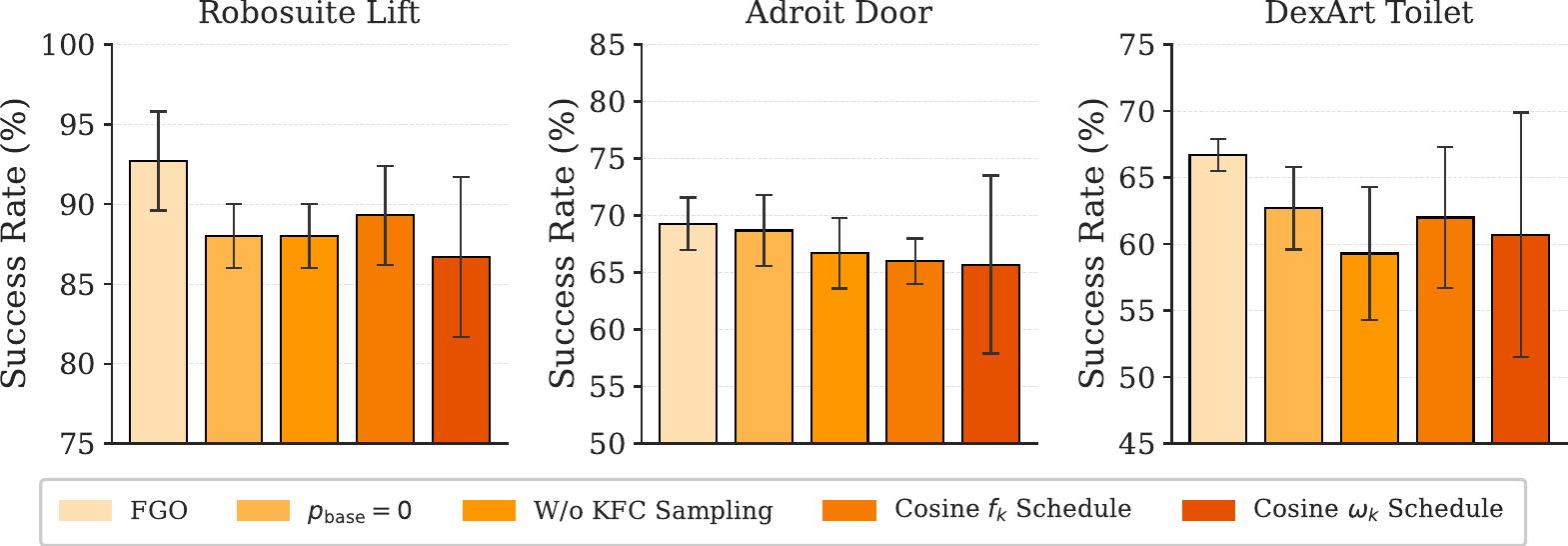}
    \end{subfigure}
    \hfill
    \begin{subfigure}[b]{0.28\textwidth}
        \centering
        \includegraphics[width=\textwidth]{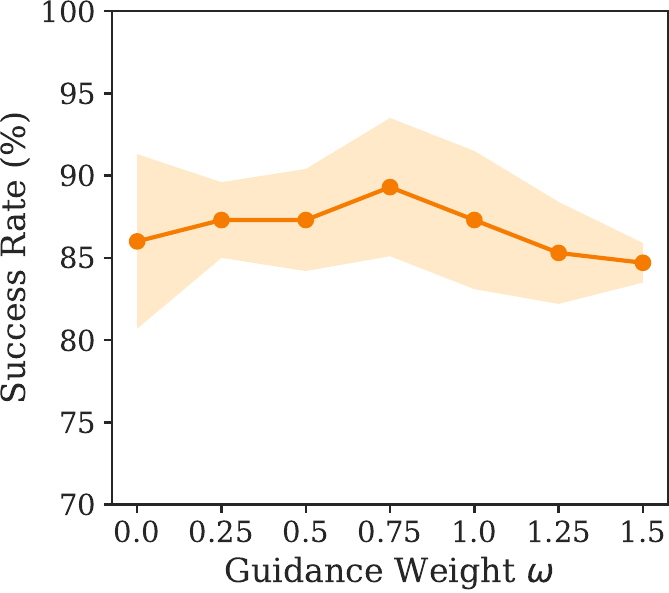}
    \end{subfigure}
    \caption{Summary of ablation experiments. (Left) Impact of individual design choices evaluated on three tasks. (Right) Performance comparison for different constant values of the guidance weight $\omega$ on the Robosuite Lift task. Here, $\omega$ remains fixed across all steps over the reverse denoising process.}
    \label{fig:ablations}
\end{figure}

We ablate our core design components to quantify their individual contributions to overall performance. The standard formulation of our approach (FGO) utilizes $p_{\text{base}} = 0.2$ and KFC sampling to regulate the cut-off frequency distribution during training, and employs linear schedules for both $f_k$ and $\omega_k$ during inference. As shown in Figure~\ref{fig:ablations}, omitting the explicit base frequency sampling ($p_{\text{base}} = 0$) degrades performance on all three tasks. This suggests that establishing a low-frequency baseline during training is critical for stabilizing the guided denoising process. Similarly, eliminating the KFC sampling strategy results in a consistent performance drop, confirming its efficacy in effectively allocating model capacity across different frequency bands. Finally, substituting the linear schedules for $f_k$ and $\omega_k$ with cosine schedules also impairs overall performance, indicating that a straightforward linear progression is effective and robust for our method.

We further provide a detailed ablation of the guidance weight $\omega$. As demonstrated in Figure~\ref{fig:ablations} (Right), an interpolation weight ($\omega \in (0, 1)$) yields superior performance compared to the extrapolation regime ($\omega > 1$). This contrasts with standard CFG \citep{HoSal2022}, which requires a large extrapolation weight to enforce strong condition adherence. We argue that this divergence stems from our frequency-based formulation. As defined in Equation~(\ref{eq:fgo_guidance}), our method blends score estimates targeting two distinct frequency manifolds. Applying an interpolation weight maintains a stable convex combination of the vector fields, thereby preserving the kinematic structure and resulting in more robust performance.

\section{Related Work}\label{related_work}

\subsection{Guidance for Diffusion and Flow Matching}
Guidance is a widely adopted technique for steering the generation process in diffusion and flow-based models \citep{HoJai2020, LipChe2022, DuDur2023, KarAit2024}. Early work \citep{DhaNic2021} introduced classifier guidance to trade off sample diversity for visual quality by steering the denoising process away from the unconditional vector field. This was later generalized by classifier-free guidance (CFG) \citep{HoSal2022}, which eliminated the need for externally trained classifiers. Subsequent research has optimized CFG by enforcing manifold constraints \citep{ChuKim2024} and approximating the unconditional score with the conditional one \citep{SadKan2024}. Beyond standard generation, guidance mechanisms have been adapted to solve complex inverse problems, such as image inpainting and super-resolution \citep{ChuKim2022, SonVah2023, PokMuc2023}, albeit with the assumption that the forward model and measurement noise are known a priori. Extending these principles to the robotics domain, early efforts applied CFG to diffusion-based policies \citep{PeaRas2023, ReuLi2023} to generate actions that adhere to the given observations or target states. Most recently, guidance techniques have been integrated into vision-language-action (VLA) models to improve temporal consistency \citep{BlaGal2025} and action coherence \citep{ParKim2025} during real-time execution.

\subsection{Frequency Modeling in Generative Models}
A novel approach for understanding and shaping the inductive biases of deep generative models is to analyze and manipulate the frequency domain. Foundational analyses of generative adversarial networks (GANs) revealed a distinct spectral bias: while generators adeptly capture low-frequency global structures, they often struggle to synthesize coherent high-frequency details \citep{DzaSha2020, SchLia2021}. This observation inspired subsequent research \citep{GalHoc2021} to design generators that operate directly within the wavelet domain, where high-frequency content can be explicitly identified and modeled. Extending spectral analysis to diffusion-based models, \citep{RisHei2022} demonstrated that high frequencies are attenuated much faster than low frequencies during the forward diffusion process. This induces a coarse-to-fine denoising process where low frequencies are reconstructed before high frequencies. To explicitly handle these spectral dynamics, later studies have sought to optimize the diffusion noise schedule either through multi-scale spatial adjustments \citep{HooHee2023} or via direct modulation in Fourier space \citep{FalPan2025, JirEar2025}. Parallel to continuous diffusion models, an emerging line of work formulates image generation as an autoregressive sequence modeling problem in the frequency domain. By transforming spatial images into quantized DCT vectors \citep{NasMen2021} or continuous frequency tokens \citep{YuLuo2025}, these methods enable the step-by-step generation of an image's spectral content. Recently, the robotics field has leveraged frequency-based techniques to develop efficient action tokenizers \citep{PerSta2025} and visuomotor policies \citep{ZhoLiu2025}.

\section{Limitations}
Our approach exhibits several limitations. First, as discussed in Section~\ref{subsec:benchmark-results}, the guidance technique introduces additional computational overhead during inference. This added latency can negatively affect kinematic responsiveness in tasks requiring high-frequency control loops. Second, we observe that the guided denoising process can occasionally generate over-smoothed action trajectories, which are detrimental to fine-grained manipulation tasks that necessitate high-precision action predictions.
\section{Conclusion}\label{sec:conclusion}
In this paper, we present Frequency Guidance Operator (FGO), a novel diffusion guidance paradigm that leverages frequency-domain inductive biases to maneuver the reverse denoising process. By training on a spectrum of low-pass-filtered action trajectories, our method enables the model to learn multi-band mappings from noise to sub-frequency data manifolds. During reverse denoising, the composite vector field progressively drives noisy samples toward the full-frequency manifold through a hierarchy of expanding sub-frequency manifolds. Extensive experiments validate that our approach achieves state-of-the-art policy performance while significantly improving the smoothness and temporal consistency of the generated action trajectories.


\bibliography{refs}

\clearpage

\appendix
\section{Derivation of $\mathbf{A}_t^{k, f}$ from $\mathbf{A}_t^k$ and $\mathbf{A}_t^0$}\label{appendix:derivation}
For a full-frequency action trajectory $\mathbf{A}_t^{0}$, its frequency-truncated counterpart $\mathbf{A}_t^{0, f}$ is defined via a low-pass filter $\mathcal{L}_f$ at cut-off frequency $f$. We can equivalently express this using the complementary high-pass filter $\mathcal{H}_f$, such that $\mathbf{A}_t^{0, f} = \mathcal{L}_f(\mathbf{A}_t^{0}) = \mathbf{A}_t^{0} - \mathcal{H}_f(\mathbf{A}_t^{0})$. Substituting this equation into the forward diffusion process, we can express the frequency-truncated noisy state $\mathbf{A}_t^{k, f}$ in terms of the full-frequency noisy state $\mathbf{A}_t^{k}$ as:
\begin{align}
    \mathbf{A}_t^{k, f} &= \sqrt{\bar{\alpha}_k}\mathbf{A}_t^{0, f} + \sqrt{1 - \bar{\alpha}_k}\boldsymbol{\epsilon}, \\
    &= \sqrt{\bar{\alpha}_k}(\mathbf{A}_t^{0} - \mathcal{H}_f(\mathbf{A}_t^{0})) + \sqrt{1 - \bar{\alpha}_k}\boldsymbol{\epsilon}, \\
    &= (\sqrt{\bar{\alpha}_k}\mathbf{A}_t^{0} + \sqrt{1 - \bar{\alpha}_k}\boldsymbol{\epsilon}) - \sqrt{\bar{\alpha}_k}\mathcal{H}_f(\mathbf{A}_t^{0}), \\
    &= \mathbf{A}_t^{k} - \sqrt{\bar{\alpha}_k}\mathcal{H}_f(\mathbf{A}_t^{0}).
\end{align}

\section{Experimental Setup}\label{appendix:experimental_setup}
Table~\ref{tab:task-configs} details the experimental configurations for all 15 tasks across 4 simulation benchmarks and one real-world platform. Within the Robosuite and MimicGen environments, the embodiment hardware is the Franka Emika Panda robot equipped with a default gripper, with visual observations obtained from dual-viewpoint point clouds. For the dexterous manipulation tasks in the Adroit and DexArt benchmarks, we utilize the Shadow Hand and Allegro Hand, respectively, constrained to single-viewpoint point cloud observations. For the real-world experimental setup, we defer further details to Appendix~\ref{appendix:real-world-experiment-details}.

\begin{table}[htbp]
\caption{Summary of task configurations. \textbf{Robot}: robotic platform used; \textbf{Objects}: total object count in the scene; \textbf{Cameras}: number of camera viewpoints; \textbf{Points}: point cloud size; \textbf{Action Dim}: degrees of freedom (DoF) in the action space; \textbf{Demos}: dataset size of expert demonstrations; \textbf{Steps}: maximum rollout horizon.}
\label{tab:task-configs}
\centering
\renewcommand\arraystretch{1.5}
\setlength{\tabcolsep}{4pt}
\scalebox{1.0}{
    \begin{tabular}{c !{\hspace{6pt}\vrule\hspace{6pt}} c c c c c c c}
    \toprule
    {\textbf{Task}} & {\textbf{Robot}} & {\textbf{Objects}} & {\textbf{Cameras}} & {\textbf{Points}} & {\textbf{Action Dim}} & {\textbf{Demos}} & {\textbf{Steps}} \\
    \midrule
    Lift & Franka & 1 & 2 & 512 & 10 & 20 & 200  \\
    Can & Franka & 1 & 2 & 1024 & 10 & 50 & 300   \\
    Square & Franka & 1 & 2 & 512 & 10 & 40 & 300  \\
    Stack & Franka & 2 & 2 & 512 & 7 & 40 & 200  \\
    \midrule
    Three Piece Assembly & Franka & 3 & 2 & 512 & 7 & 50 & 500  \\
    Stack Three & Franka & 3 & 2 & 512 & 7 & 50 & 400 \\
    \midrule
    Hammer & Shadow & 1 & 1 & 512 & 26 & 10 & 300 \\
    Door & Shadow & 1 & 1 & 512 & 28 & 10 & 300 \\
    Pen & Shadow & 2 & 1 & 512 & 24 & 10 & 300 \\
    \midrule
    Laptop & Allegro & 1 & 1 & 1024 & 22 & 100 & 50 \\
    Toilet & Allegro & 1 & 1 & 1024 & 22 & 100 & 50  \\
    Faucet & Allegro & 1 & 1 & 1024 & 22 & 100 & 50  \\
    Bucket & Allegro & 1 & 1 & 1024 & 22 & 100 & 50  \\
    \midrule
    Cup & xArm & 2 & 1 & 1024 & 10 & 25 & 50 \\
    Mouse & xArm & 1 & 1 & 1024 & 10 & 25 & 50 \\
    \bottomrule
    \end{tabular}
}
\end{table}
\begin{figure}[htbp]
    \centering
    \begin{subfigure}[b]{0.3\textwidth}
        \centering
        \includegraphics[width=\textwidth]{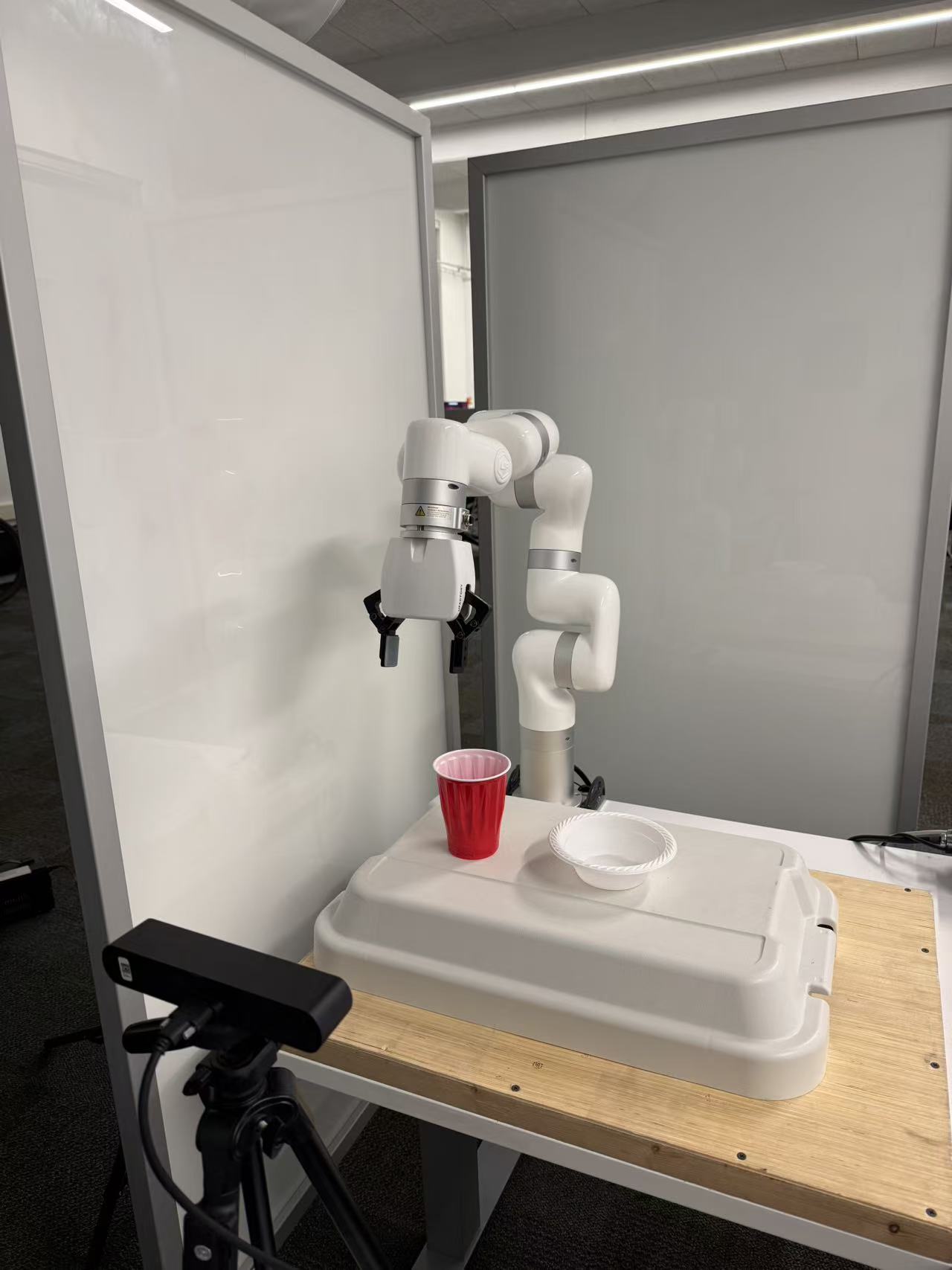}
    \end{subfigure}
    \hfill
    \begin{subfigure}[b]{0.3\textwidth}
        \centering
        \includegraphics[width=\textwidth]{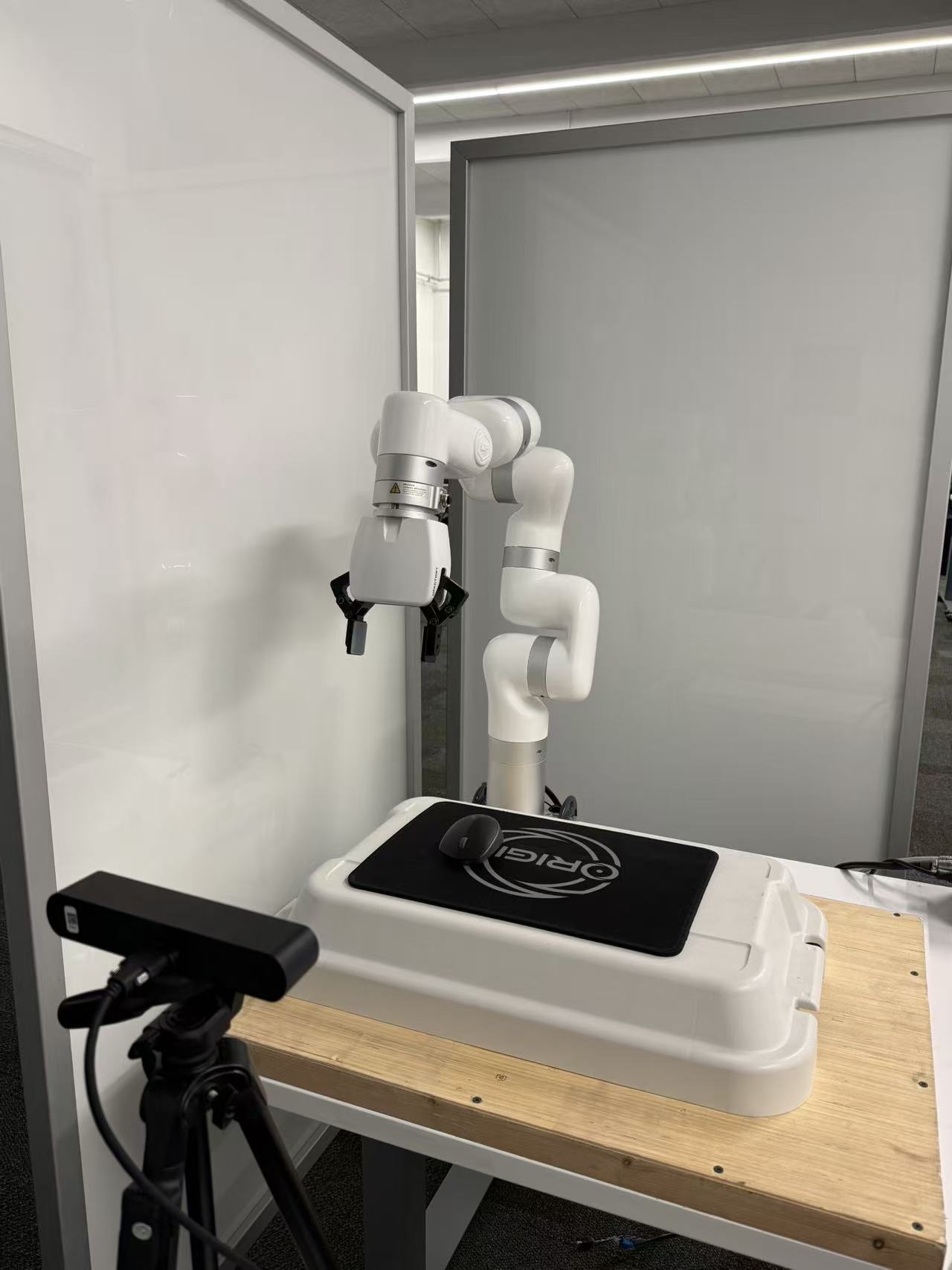}
    \end{subfigure}
    \hfill
    \begin{subfigure}[b]{0.3\textwidth}
        \centering
        \includegraphics[width=\textwidth]{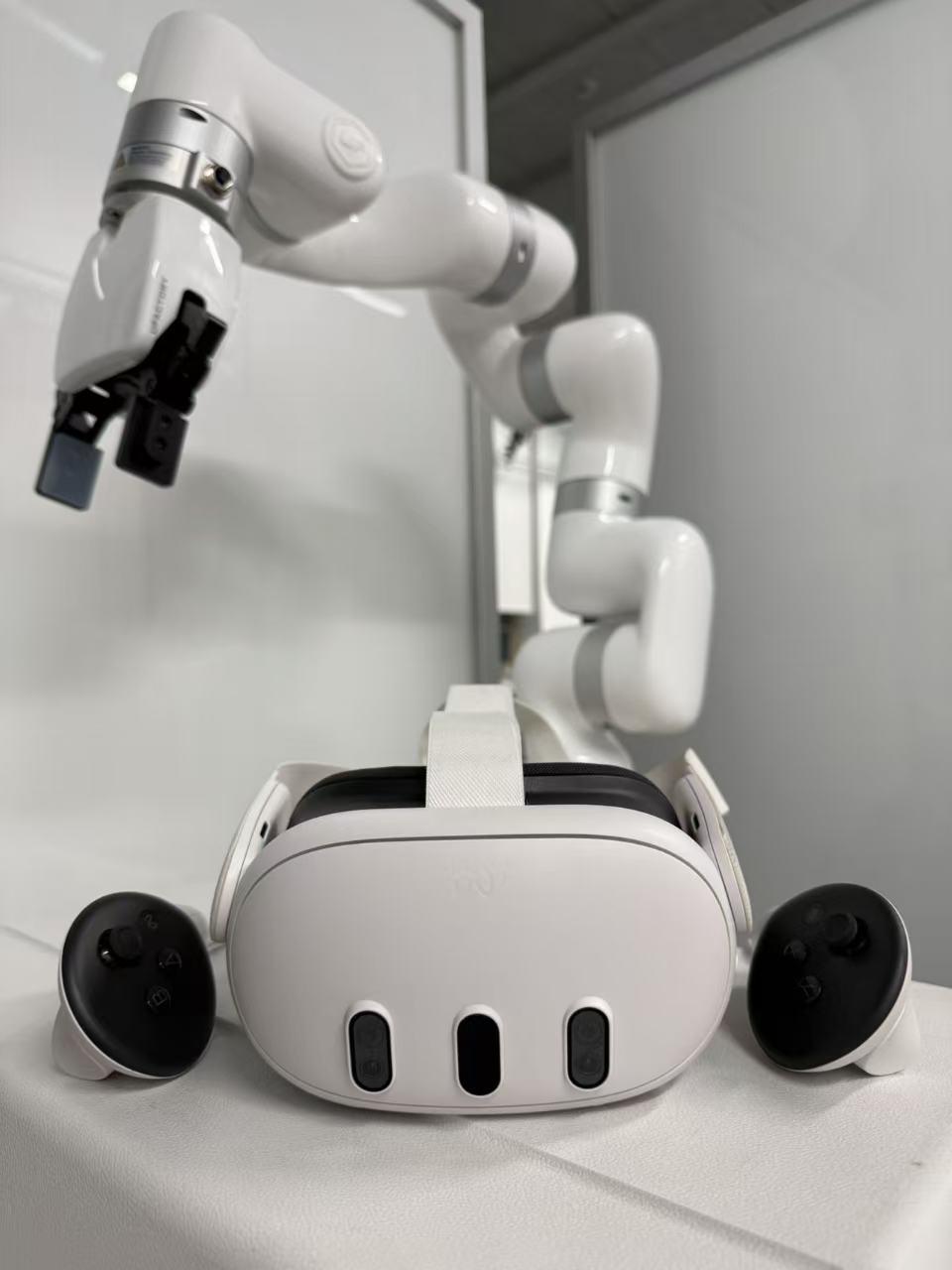}
    \end{subfigure}
    \caption{Hardware for real-world experiments. (Left) Physical workspace setup for the Cup task. (Middle) Physical workspace setup for the Mouse task. (Right) The teleoperation apparatus used for data collection.}
    \label{fig:physical-hardware}
\end{figure}

\section{Real-World Experiment Details}\label{appendix:real-world-experiment-details}

\subsection{Physical Hardware}
The real-world experiments are conducted in a tabletop workspace equipped with an xArm robotic manipulator and a ZED 2 stereo camera, as shown in Figure~\ref{fig:physical-hardware}.

\subsection{Observation and Action Space}
The observation space comprises robot proprioception from the xArm manipulator (arm joint positions, end-effector poses, and gripper states) and a third-person point cloud captured by the ZED camera. The raw visual data from the camera is first downsampled via a pixel stride, flattened, and filtered to remove non-finite points. The remaining valid points are then spatially cropped to a predefined 3D bounding box over the workspace. Finally, Farthest Point Sampling (FPS) \citep{QiYi2017} is applied to extract a fixed-size representation of 1,024 points. The corresponding action space is defined by the absolute robot end-effector poses and the gripper joint positions.

\subsection{Data Collection}
As illustrated in Figure~\ref{fig:physical-hardware} (Right), expert demonstrations are collected using a Meta Quest 3 headset and its spatial controllers to teleoperate the xArm manipulator. The VR hand controllers provide real-time 6-DoF pose tracking, which is kinematically translated into target poses for the robot's end-effector, while analog trigger inputs control the gripper opening and closing. During the teleoperation process, the physical state of the robot, the visual point cloud observations, and the expert actions are logged asynchronously to prevent control latency. Following the online collection phase, an offline post-processing pipeline aligns the multimodal data streams using their respective timestamps. This decoupled logging strategy eliminates the computational overhead of real-time synchronization and thus ensures a highly responsive teleoperation interface.

\subsection{Evaluation}
During evaluation, the pretrained policies are deployed on the xArm manipulator over 25 independent trials. To test the robustness of the policies, object initial poses are randomized within predefined bounds at the start of each episode. At every control step, the policy maps the multimodal observations into a sequence of predicted actions, which are then dispatched to the robot's low-level controller. An episode ends upon success, an unsafe collision, or a timeout after a strict limit of 50 time steps.

\section{Implementation Details}
\subsection{Model Architecture}
While the proposed FGO framework is fundamentally agnostic to the underlying diffusion-based policy network, we empirically validate our method by integrating it into the DP3 architecture \citep{ZeZha2024}. Concretely, we introduce a lightweight, MLP-based encoder to condition the model on the cut-off frequency, leaving the core network unmodified. This frequency encoder shares the exact architectural design as the diffusion step encoder utilized in DP3.

\subsection{Training and Inference}
Model training and inference across all the evaluated algorithms are executed on a single NVIDIA RTX 4090 GPU. During training, we utilize a batch size of 512 for the Robosuite and MimicGen benchmarks, and a batch size of 128 for the Adroit, DexArt, and real-world benchmarks. All models are trained for a total of 3,000 epochs and evaluated at intervals of 600 epochs.

\subsection{Hyperparameters}
The hyperparameter configurations for our full method (FGO) are summarized in Table~\ref{tab:hyperparameters}. Empirically, FGO yields robust performance with $f_{\text{base}} = 3$ and $p_{\text{base}} = 0.2$ across the majority of tasks, with the notable exception of the DexArt Laptop and Toilet tasks, where $f_{\text{base}} = 0$ achieves optimal results. Furthermore, we observe that the optimal values of $\beta$ for KFC sampling vary depending on the specific task. This variation is expected, as the coupling between spectral frequencies and noise levels during the diffusion denoising process differs significantly across tasks. Finally, employing linear schedules for $f_k$ and $\omega_k$ yields the best performance for all tasks.

\begin{table}[htbp]
\caption{Summary of hyperparameter configurations.}
\label{tab:hyperparameters}
\centering
\renewcommand\arraystretch{1.5}
\setlength{\tabcolsep}{4pt}
\scalebox{1.0}{
    \begin{tabular}{c !{\hspace{6pt}\vrule\hspace{6pt}} c c c !{\vrule} c c}
    \toprule
    \multirow{3}{*}{\raisebox{1.3\height}{\textbf{Task}}} & \multicolumn{3}{c}{\textbf{Training}} & \multicolumn{2}{c}{\textbf{Inference}} \\
    \cmidrule(r){2-4}
    \cmidrule(l){5-6}
     & {$f_{\text{base}}$} & {$p_{\text{base}}$} & {$\beta$} & {$f_k$ Schedule} & {$\omega_k$ Schedule} \\
    \midrule
    Lift & 3 & 0.2 & 0.5 & Linear & Linear \\
    Stack & 3 & 0.2 & 0.5 & Linear & Linear \\
    Can & 3 & 0.2 & 0.5 & Linear & Linear \\
    Square & 3 & 0.2 & 0.5 & Linear & Linear \\
    \midrule
    Three Piece Assembly & 3 & 0.2 & 0.5 & Linear & Linear \\
    Stack Three & 3 & 0.2 & 0.5 & Linear & Linear \\
    \midrule
    Hammer & 3 & 0.2 & 0.0 & Linear & Linear \\
    Door & 3 & 0.2 & 0.5 & Linear & Linear \\
    Pen & 3 & 0.2 & 0.0 & Linear & Linear \\
    \midrule
    Laptop & 0 & 0.2 & 1.0 & Linear & Linear \\
    Toilet & 0 & 0.2 & 1.0 & Linear & Linear \\
    Faucet & 3 & 0.2 & 0.0 & Linear & Linear \\
    Bucket & 3 & 0.2 & 1.0 & Linear & Linear \\
    \midrule
    Cup & 3 & 0.2 & 0.5 & Linear & Linear \\
    Mouse & 3 & 0.2 & 0.5 & Linear & Linear \\
    \bottomrule
    \end{tabular}
}
\end{table}
\begin{table}[htbp]
\caption{Comparison of success rates (\%) against alternative guidance methods on the Robosuite \citep{ZhuWon2020} and MimicGen \citep{ManNas2023} benchmarks.}
\label{tab:supplementary-success-1}
\centering
\renewcommand\arraystretch{1.5}
\setlength{\tabcolsep}{4pt}
\scalebox{0.7}{
    \begin{tabular}{l !{\hspace{6pt}\vrule\hspace{6pt}} *{4}{S[table-format=3.1(1.1)]} !{\vrule} *{2}{S[table-format=3.1(1.1)]} !{\vrule} S[table-format=1.0]}
    \toprule
    \multirow{2}{*}{\textbf{Algorithms / Tasks}} & \multicolumn{4}{c}{\textbf{Robosuite}} & \multicolumn{2}{c}{\textbf{MimicGen}} & {\multirow{2}{*}{\textbf{Average}}} \\
    \cmidrule(r){2-5}
    \cmidrule(lr){6-7}
     & {Lift} & {Stack} & {Can} & {Square} & {Three Piece Assembly} & {Stack Three} & {} \\
    \midrule
    CFG \citep{HoSal2022} + DP3 \citep{ZeZha2024} & 91.3 \pm 3.1 & 66.7 \pm 5.8 & 59.3 \pm 3.1 & 29.3 \pm 4.2 & 24.7 \pm 4.2 & 17.3 \pm 1.2 & 48.1 \\
    ACG \citep{ParKim2025} + DiT-Policy \citep{DasMee2025} & 88.0 \pm 4.0 & 26.7 \pm 9.8 & 34.0 \pm 7.2 & 4.0 \pm 3.5 & 0.0 \pm 0.0 & 0.0 \pm 0.0 & 25.5 \\
    \bottomrule
    \end{tabular}
}
\end{table}

\begin{table}[htbp]
\caption{Comparison of success rates (\%) against alternative guidance methods on the Adroit \citep{RajKum2017} and DexArt \citep{BaoXu2023} benchmarks.}
\label{tab:supplementary-success-2}
\centering
\renewcommand\arraystretch{1.5}
\setlength{\tabcolsep}{4pt}
\scalebox{0.7}{
    \begin{tabular}{l !{\hspace{6pt}\vrule\hspace{6pt}} *{3}{S[table-format=3.1(1.1)]} !{\vrule} *{4}{S[table-format=3.1(1.1)]} !{\vrule} S[table-format=1.0]}
    \toprule
    \multirow{2}{*}{\textbf{Algorithms / Tasks}} & \multicolumn{3}{c}{\textbf{Adroit}} & \multicolumn{4}{c}{\textbf{DexArt}} & {\multirow{2}{*}{\textbf{Average}}} \\
    \cmidrule(r){2-4}
    \cmidrule(lr){5-8}
     & {Hammer} & {Door} & {Pen} & {Laptop} & {Toilet} & {Faucet} & {Bucket} & {} \\
    \midrule
    CFG \citep{HoSal2022} + DP3 \citep{ZeZha2024} & 100\pm0.0 & 62.3\pm8.5 & 47.7\pm2.5 & 71.3\pm6.1 & 59.3\pm7.0 & 22.7\pm2.3 & 22.7\pm5.0 & 55.1 \\
    ACG \citep{ParKim2025} + DiT-Policy \citep{DasMee2025} & 92.7\pm9.0 & 57.3\pm9.0 & 47.7\pm2.1 & 68.7\pm9.0 & 60.7\pm8.1 & 22.7\pm4.2 & 20.7\pm2.3 & 52.9 \\
    \bottomrule
    \end{tabular}
}
\end{table}

\begin{table}[h!]
\caption{Comparison of success rates (\%) against alternative action smoothing methods on the Robosuite \citep{ZhuWon2020} and MimicGen \citep{ManNas2023} benchmarks.}
\label{tab:supplementary-success-3}
\centering
\renewcommand\arraystretch{1.5}
\setlength{\tabcolsep}{4pt}
\scalebox{0.68}{
    \begin{tabular}{l !{\hspace{6pt}\vrule\hspace{6pt}} *{4}{S[table-format=3.1(1.1)]} !{\vrule} *{2}{S[table-format=3.1(1.1)]} !{\vrule} S[table-format=1.0]}
    \toprule
    \multirow{2}{*}{\textbf{Algorithms / Tasks}} & \multicolumn{4}{c}{\textbf{Robosuite}} & \multicolumn{2}{c}{\textbf{MimicGen}} & {\multirow{2}{*}{\textbf{Average}}} \\
    \cmidrule(r){2-5}
    \cmidrule(lr){6-7}
     & {Lift} & {Stack} & {Can} & {Square} & {Three Piece Assembly} & {Stack Three} & {} \\
    \midrule
    Low-Pass Filter ($f = 4$) + DP3 \citep{ZeZha2024} & 86.0 \pm 5.3 & 66.0 \pm 4.0 & 62.7 \pm 4.6 & 20.7 \pm 4.2 & 26.0 \pm 6.0 & 14.7 \pm 2.3 & 46.0 \\
    Low-Pass Filter ($f = 8$) + DP3 \citep{ZeZha2024} & 88.0 \pm 3.5 & 69.3 \pm 5.0 & 61.3 \pm 2.3 & 20.7 \pm 3.1 & 33.3 \pm 8.3 & 17.3 \pm 3.1 & 48.3 \\
    Temporal Ensembling \citep{ZhaKum2023} + DP3 \citep{ZeZha2024} & 85.3 \pm 4.6 & 69.3 \pm 1.2 & 62.0 \pm 2.0 & 10.7 \pm 6.4 & 33.3 \pm 1.2 & 25.3 \pm 3.1 & 47.7 \\
    \bottomrule
    \end{tabular}
}
\end{table}

\begin{table}[h!]
    \centering
    \caption{Comparison of ATV and JerkRMS against alternative action smoothing methods on the Robosuite Can task.}
    \label{tab:supplementary-atv-jerkrms}
    \renewcommand\arraystretch{1.5}
    \setlength{\tabcolsep}{4pt}
    \scalebox{0.8}{
        \begin{tabular}{l !{\hspace{6pt}\vrule\hspace{6pt}} *{2}{S[table-format=2.2(1.2)]}}
        \toprule
        {Algorithms} & {ATV $\downarrow$ ($\times 10^{-3}$ rad/s)} & {JerkRMS $\downarrow$ (rad/s$^3$)} \\
        \midrule
        Low-Pass Filter ($f = 4$) + DP3 \citep{ZeZha2024} & 14.74 \pm 0.10 & 35.21 \pm 0.91 \\
        Low-Pass Filter ($f = 8$) + DP3 \citep{ZeZha2024} & 14.79 \pm 0.12 & 40.43 \pm 0.88 \\
        Temporal Ensembling \citep{ZhaKum2023} + DP3 \citep{ZeZha2024} & 14.73 \pm 0.06 & 44.91 \pm 1.35 \\
        \bottomrule
        \end{tabular}
    }
\end{table}

\section{Supplementary Results}\label{appendix:supplementary-results}
For a comprehensive comparison against existing guidance techniques, we benchmark two additional guidance methods on the simulation environments described in Section~\ref{subsec:baselines-simulation-benchmarks}. These baselines are constructed by incorporating classifier-free guidance (CFG) \citep{HoSal2022} into the DP3 architecture \citep{ZeZha2024} as proposed by \citep{PeaRas2023}, and by integrating action coherence guidance (ACG) \citep{ParKim2025} into DiT-Policy \citep{DasMee2025, ZhuYu2025}. As shown in Tables~\ref{tab:supplementary-success-1} and \ref{tab:supplementary-success-2}, the introduction of CFG to DP3 negatively impacts performance on 8 tasks, results in negligible change on 1 task, and improves success rates on the remaining 4 tasks. This phenomenon aligns with the findings from \citep{PeaRas2023}, which hypothesized that CFG over-amplifies observation-specific behaviors and prompts the policy to predict atypical, low-probability actions rather than robust, high-likelihood trajectories. When ACG is coupled with DiT-Policy, it leads to broad performance degradation across all but the DexArt Faucet and Bucket tasks. This failure mode likely stems from the over-smoothing effect of ACG; by enforcing rigid intra-chunk action coherence, the approach inevitably corrupts fine-grained adjustments that are necessary for task completion.

We also compare against two alternative action smoothing methods. The first method directly applies a low-pass filter with a cut-off frequency $f$ to the predicted action trajectories. The second is the temporal ensembling technique \citep{ZhaKum2023}, which computes a weighted sum of actions predicted at different time steps to synthesize a single-step action. As shown in Tables~\ref{tab:supplementary-success-3} and \ref{tab:supplementary-atv-jerkrms}, while both methods enable the policy to generate smoother actions (lower ATV and JerkRMS), they concurrently degrade success rates on the majority of tasks. For low-pass filtering, this performance drop occurs because the filter simultaneously removes fine-grained details alongside high-frequency noise within the action trajectories. For temporal ensembling, the averaging operation forces multi-modal action predictions to collapse into a single modality, which disrupts the underlying kinematic structure of the trajectory and induces contradictory control signals.
\section{Frequency Analysis}
\begin{figure}[htbp]
    \centering
    \begin{subfigure}[b]{0.45\textwidth}
        \centering
        \includegraphics[width=\textwidth]{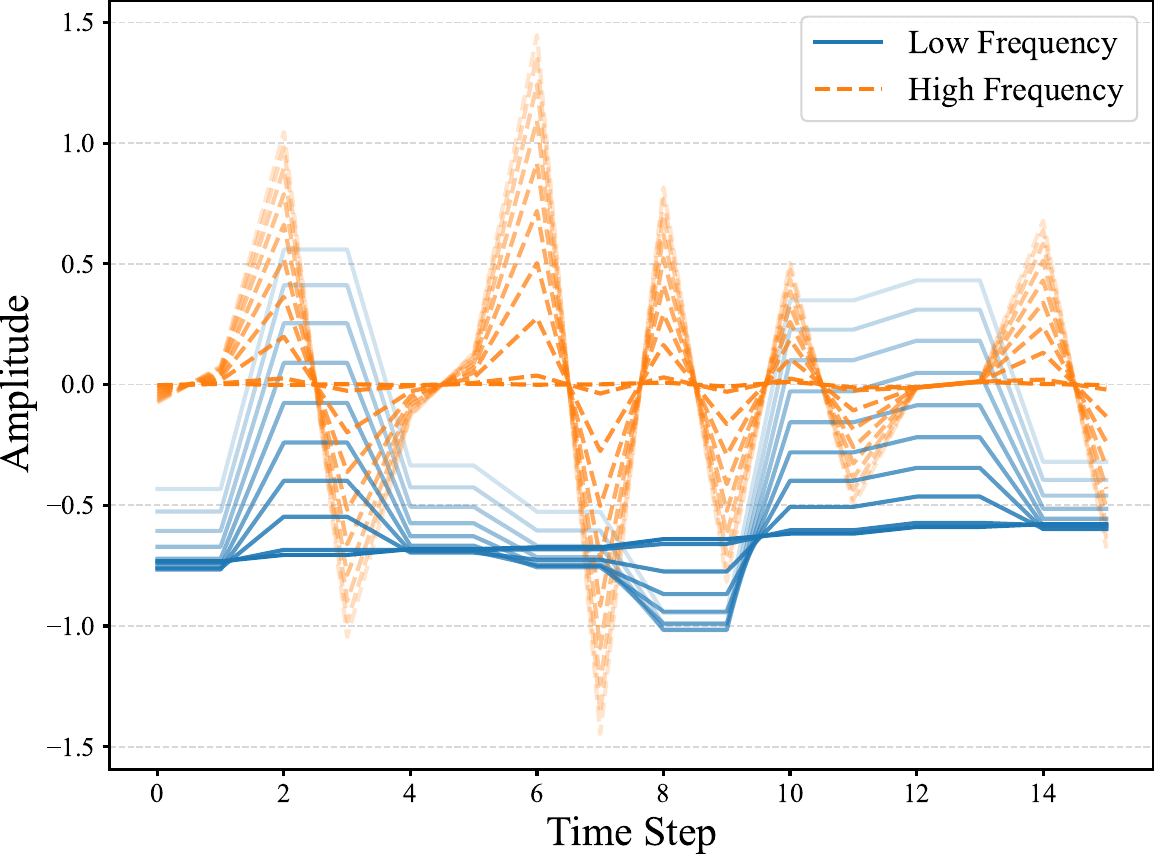}
    \end{subfigure}
    \hfill
    \begin{subfigure}[b]{0.45\textwidth}
        \centering
        \includegraphics[width=\textwidth]{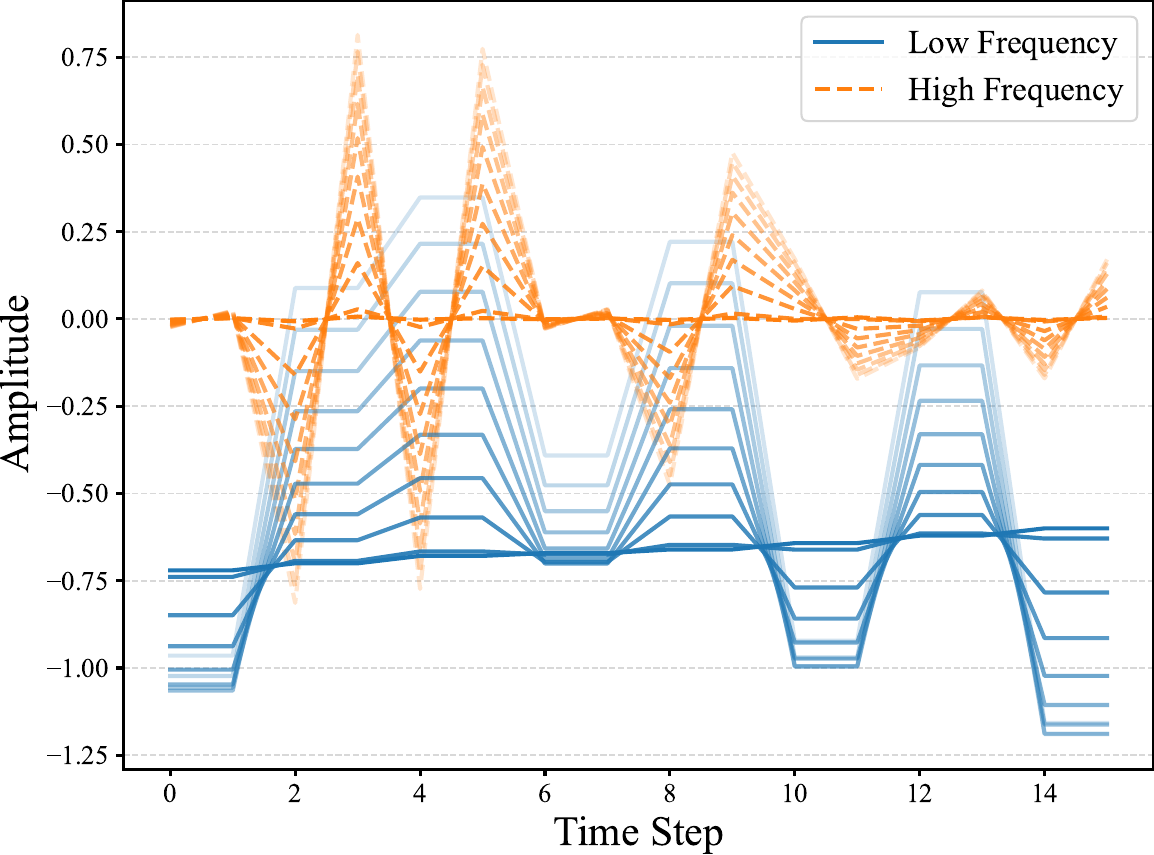}
    \end{subfigure}
    \caption{Evolution of low-frequency and high-frequency action components during the reverse denoising process. We compare trajectories generated by DP3 \citep{ZeZha2024} (left) against our method (right). Color intensity increases (light to dark) as the diffusion step decreases from $k=K$ to $k=0$.}
    \label{fig:trajectory-evolution}
\end{figure}

In this section, we analyze the frequency characteristics of the generated action trajectories throughout the reverse denoising process. Concretely, at each diffusion step, we apply the discrete Haar wavelet transform \citep{Haa1909, StaFal2003} to decompose the full-frequency trajectory into low-frequency and high-frequency components. As visualized in Figure~\ref{fig:trajectory-evolution}, the low-frequency components are progressively refined into a smooth trajectory, while the high-frequency components gradually attenuate and converge to zero by the end of the denoising process. Compared to the DP3 baseline, our approach generates trajectories with notably less high-frequency variance during intermediate steps, indicating that the high-frequency noise is effectively suppressed.

\end{document}